\PassOptionsToPackage{table}{xcolor}
\documentclass[11pt]{article}

\usepackage[final]{acl}

\usepackage{times}
\usepackage{latexsym}

\usepackage[T1]{fontenc}
\usepackage[utf8]{inputenc}

\usepackage{microtype}

\usepackage{inconsolata}

\usepackage{graphicx}
\usepackage{booktabs}
\usepackage{multirow}
\usepackage{xcolor}
\usepackage{enumitem}
\usepackage{amssymb}
\usepackage{makecell}

\title{CultureVidBench: Benchmarking Cultural Understanding in\\Text-to-Video Generation}

\author{
 \textbf{Xianjing Han\textsuperscript{1}},
 \textbf{Yuhan Su\textsuperscript{2}},
 \textbf{Yang Deng\textsuperscript{3}},
 \textbf{Dong Ma\textsuperscript{4}},
 \textbf{Wee Peng Tay\textsuperscript{1}},
 \textbf{Bin Zhu\textsuperscript{3} \thanks{Corresponding author and project lead.}}
\\
 \textsuperscript{1} Nanyang Technological University \textsuperscript{2} Centrale Supélec \\
 \textsuperscript{3} Singapore Management University
 \textsuperscript{4} University of Cambridge
\\
 \small{
   \textbf{Correspondence:} \href{mailto:binzhu@smu.edu.sg}{binzhu@smu.edu.sg}
 }
 \\
\small{
\url{https://hanxjing.github.io/CultureVidBench/}
}
}
\begin{document}
\maketitle

\begin{abstract}
Text-to-video (T2V) generation models have advanced rapidly, yet their ability to represent diverse cultural contexts remains underexplored. Existing benchmarks mainly focus on perceptual quality, physical plausibility, and text-video alignment, but do not directly assess whether generated videos capture culturally specific objects, actions, rituals, visible text, or audio cues.
We introduce CultureVidBench, a comprehensive benchmark for evaluating cultural understanding in T2V generation. CultureVidBench contains 1,000 curated prompts covering 12 countries, 6 continents, 8 cultural regions, and 14 cultural aspects organized into three categories: material culture, social practice \& performance, and ritual \& ceremony. Designed specifically for video generation, CultureVidBench emphasizes dynamic and multimodal cultural representation, including social interactions, ritual procedure, and culturally appropriate visible text and audio. We evaluate seven representative T2V models through human user studies and MLLM-based automatic assessment across cultural faithfulness, multimodal cultural rendering, semantic adherence, and perceptual quality.
Results show that although current models achieve strong semantic adherence and visual quality, they often fail to faithfully capture fine-grained cultural details, particularly for underrepresented regions, rituals, and multimodal cultural cues.
\end{abstract}

\section{Introduction}
\begin{figure*}[t]
	\centering
	\includegraphics[width=0.94\textwidth]{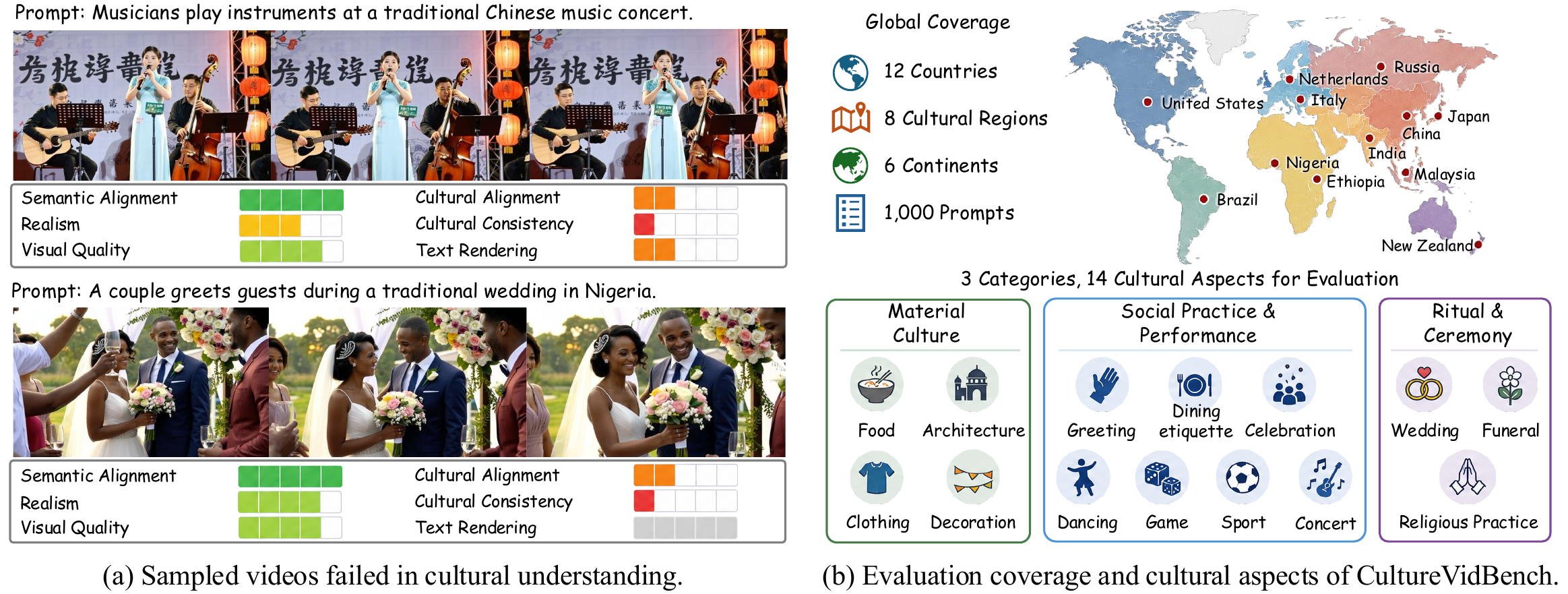}
    \vspace{-2mm}
	\caption{Cultural failures in T2V generation and overview of CultureVidBench. (a) Examples of semantically aligned and visually plausible generated videos with cultural inaccuracies. (b) CultureVidBench cultural coverage.}
	\label{Fig: intro}
    \vspace{-3mm}
\end{figure*}

Recent advances in text-to-video (T2V) generation have enabled models to produce increasingly realistic, temporally coherent, and semantically aligned videos from natural language prompts~\citep{wan2025, Veo2025, happyhorse2026}. These models are rapidly adopted to support a growing range of applications, including filmmaking, social media content creation, advertising, and education~\citep{zheng2025vbench,chen2026t2vworldbench}. %As generated videos become more widely created and consumed by users across diverse regions and societies, an important question emerges: can T2V models faithfully understand the cultural contexts described in user prompts?
As generated videos become widely consumed across different societies, an important question emerges: \textit{can T2V models faithfully understand and represent cultural contexts?}

% their ability to understand and faithfully represent cultural contexts has become a critical concern~\citep{geertz2017interpretation}.

% Culture fundamentally shapes real-world videos, from clothing and architecture to rituals, social interactions, human activities, and collective behaviors~\citep{rege2025cure}. Therefore, high-quality and authentic video generation requires not only semantic correctness, but also culturally grounded generation. A video may be semantically aligned with the prompt while still being culturally inaccurate. 

Culture is deeply embedded in real-world videos. It shapes not only visible objects such as clothing, food, architecture, and decorations, but also human actions, social interactions, ritual procedures, performances, music, and symbolic practices~\citep{geertz2017interpretation}. Therefore, cultural understanding in T2V generation cannot be limited to general prompt following or visual realism. 
%A video may appear fluent, aesthetically pleasing, and semantically related to the prompt, yet still misrepresent the target culture. As shown in Figure~\ref{Fig: intro} (a), a generated video may appear visually plausible and semantically related to the prompt while still misrepresenting the intended cultural context. For a traditional Chinese music concert, the model may depict people performing on a stage but introduce culturally inconsistent musical elements or visible text. For a traditional Nigerian wedding, the model may capture the coarse event structure of a couple greeting guests, yet render generic or non-target wedding attire, symbols, and scene composition.
As shown in Figure~\ref{Fig: intro}(a), a generated video may appear visually plausible and semantically related to the prompt while still misrepresenting the intended culture. For example, a model may depict a traditional Chinese music performance using culturally inconsistent instruments or text, or generate a Nigerian wedding scene with generic attire and ceremonial elements.
% a model prompted to generate a traditional Chinese music concert may instead produce generic Western musical instruments, while a generated traditional Nigerian wedding may include Western wedding attire.
Such videos may score highly for visual quality and coarse semantic alignment, but they remain culturally inaccurate, misleading, or inappropriate for viewers in the target culture. These failures suggest that cultural understanding is a central requirement for trustworthy T2V generation, rather than a secondary aesthetic preference.
% Such culturally inconsistent content may appear unrealistic, inappropriate, or even misleading to viewers from the target cultural background.
% These failures suggest that cultural understanding is not an auxiliary property of T2V systems, but a central requirement for faithful and trustworthy video generation.

Despite its importance, cultural understanding remains largely underexplored in existing T2V evaluation. Current benchmarks primarily assess perceptual quality~\citep{huang2024vbench}, motion consistency~\citep{liao2024evaluation}, physical plausibility~\citep{gu2025phyworldbench}, and text-video semantic alignment~\citep{sun2025t2v}. While these dimensions are essential, they do not directly evaluate whether a generated video correctly captures culturally specific objects, behaviors, rituals, visible text, or audio cues.
Recent cultural benchmarks for text-to-image generation have begun to evaluate cultural competence in static images~\citep{kannen2024beyond,shi2025culture,malakouti2026culture}, but they cannot fully capture the video-specific nature of culture. 
%Compared with images, videos require models to represent culture as it unfolds over time: people greet, dance, dine, perform ceremonies, participate in weddings, and interact according to culturally specific norms. Moreover, cultural meaning may be expressed through multiple modalities, including visual appearance, written language, music, chanting, and ambient sound. These properties make cultural understanding in T2V generation a distinct and more challenging problem.
Unlike images, videos require models to represent culturally grounded actions and interactions unfolding over time, often involving multiple modalities such as written language, music, chanting, and ambient sound.

% while largely overlooking cultural faithfulness and culturally grounded generation.
% Recent text-to-image benchmarks have explored cultural competence~\cite{kannen2024beyond,shi2025culture,malakouti2026culture}, but they mainly evaluate static images of cultural elements associated with countries. However, culture in video is often expressed through dynamic and multimodal practices, including culturally grounded actions, social interactions, rituals, visible text, and audio cues. These aspects are central to video generation but cannot be fully captured by image-based evaluation protocols.

To address this gap, we introduce CultureVidBench, a comprehensive benchmark for evaluating cultural understanding in text-to-video generation. As shown in Figure~\ref{Fig: intro} (b), CultureVidBench covers 12 countries spanning 6 continents and 8 world cultural regions, and organizes cultural content into three major categories: material culture, social practice \& performance, and ritual \& ceremony. These categories further span 14 cultural aspects, such as food, architecture, greetings, games, weddings, and religious rituals. In total, CultureVidBench contains 1,000 curated prompts constructed from culturally grounded elements and activities in CulturalAtlas~\citep{mosaica2024culturalatlas} and Wikipedia~\citep{wikipedia}.
Particularly, unlike benchmarks that focus primarily on static cultural artifacts, CultureVidBench emphasizes dynamic and multimodal cultural representation, including culturally specific actions, social interactions, procedural rituals, visible text, and audio cues. 

We further design a comprehensive evaluation framework for cultural video generation across four complementary dimensions: cultural faithfulness, multimodal cultural rendering, semantic adherence, and perceptual quality. Cultural faithfulness assesses whether the target cultural elements, practices, and contexts are faithfully represented, while also checking whether the video avoids introducing culturally inconsistent artifacts or symbols from unrelated contexts. Multimodal cultural rendering evaluates whether visible text and audio are appropriate for the target cultural scenario. Semantic adherence and perceptual quality are included to distinguish cultural failures from general prompt-following or low-quality generation failures. 
We evaluate seven representative T2V models using both human studies and MLLM-based automatic evaluation. To improve interpretability, we introduce a target-explicit evaluation protocol that explicitly identifies expected cultural elements from prompts before evaluating video evidence.

Experiments reveal several important findings: 1) While current T2V models often achieve strong semantic adherence and visual quality, they still struggle to faithfully capture culturally specific details. 2) This limitation is more pronounced in underrepresented cultural regions, where models perform substantially worse than in high-resource cultures. 3) Material culture is generally easier to generate than social practices and rituals, suggesting that current models are better at rendering cultural objects than culturally grounded actions and procedures. 4) Multimodal cultural rendering remains particularly challenging, especially for culturally appropriate visible text and audio.

Our contributions can be summarized as follows:
\begin{itemize}[leftmargin=*, itemsep=0pt, topsep=2pt, parsep=0pt, partopsep=0pt]
%\item We introduce CultureVidBench, the first benchmark for evaluating cultural understanding in text-to-video generation across diverse countries, cultural regions and cultural categories.
\item We introduce CultureVidBench, a benchmark for evaluating cultural understanding in text-to-video generation across diverse countries, cultural regions and cultural categories.
\item We propose a comprehensive cultural evaluation framework covering cultural faithfulness, multimodal cultural rendering, semantic adherence, and perceptual quality.%, combining both human user studies and structured MLLM-based automatic assessment.
\item We benchmark seven SOTA T2V models and reveal systematic cultural disparities, showing that current models remain biased toward better-represented cultures and struggle with culturally specific practices, rituals, and multimodal cues.
\end{itemize}

\section{Related Work}

\textbf{Text-to-Video Generation Benchmarks.}
Recent advances in T2V generation have highlighted the need for systematic and reliable evaluation~\citep{sun2025t2v,huang2025vbench++}. Early benchmarks, such as VBench~\citep{huang2024vbench} and EvalCrafter~\citep{liu2024evalcrafter}, assess T2V models from multiple general aspects, including visual quality, temporal consistency, motion quality, and text-video alignment. Beyond general quality assessment, recent benchmarks have examined more specific model capabilities. For example, DEVIL~\citep{liao2024evaluation} focuses on dynamic properties in generated videos, and OSCBench~\citep{han2026oscbench} evaluates object state change reasoning. Other benchmarks, including VideoPhy~\citep{bansal2024videophy}, PhyGenBench~\citep{meng2024towards}, and PhyWorldBench~\citep{gu2025phyworldbench}, evaluate whether generated videos follow physical commonsense and plausible world dynamics. RoboTrustBench~\citep{li2026robotrustbench} further evaluates video world models for robotic manipulation under normal, constraint-sensitive, counterfactual, and adversarial instructions. Despite these advances, existing T2V benchmarks still underexplore cultural understanding, which is essential for generating videos that are faithful and appropriate across diverse social contexts. To address this gap, we introduce CultureVidBench, a benchmark specifically designed for cultural evaluation in T2V generation.

\noindent\textbf{Cultural Benchmarks.}
As generative models and vision-language models (VLMs) are increasingly deployed worldwide, evaluating their ability to represent diverse cultures has become an important research direction. Existing cultural benchmarks have studied this problem across several domains~\citep{bhatt-diaz-2024-extrinsic,chen2026arteculture}. In text-to-image generation, benchmarks such as CUBE-1K~\citep{kannen2024beyond}, CULTIVate~\citep{malakouti2026culture}, ViSAGe~\citep{jha2024visage}, CultureBench~\citep{shi2025culture}, CulturalFrames~\citep{nayak2025culturalframes}, and CuRe~\citep{rege2025cure} evaluate whether generated images faithfully reflect cultural concepts, artifacts, and social contexts. In vision-language understanding, CulturalVQA~\citep{nayak2024benchmarking}, CulturalGround~\citep{nyandwi2025grounding}, CURVE~\citep{Singh_2026_CVPR}, and VideoNorms~\citep{varimalla2025videonorms} examine the cultural knowledge and grounding ability of VLMs. However, T2V generation poses distinct challenges for cultural evaluation. Unlike static images, videos require models to represent culture through temporal actions, social interactions, rituals, and multimodal cues such as visible text and audio. Therefore, protocols designed for T2I generation or VLM understanding cannot fully capture video-specific cultural challenges, especially those involving human actions, ritual procedures, and audio cultural expression. CultureVidBench addresses this gap by providing a dedicated benchmark for culturally grounded T2V generation.

\begin{figure*}[t]
	\centering
	\includegraphics[width=0.97\textwidth]{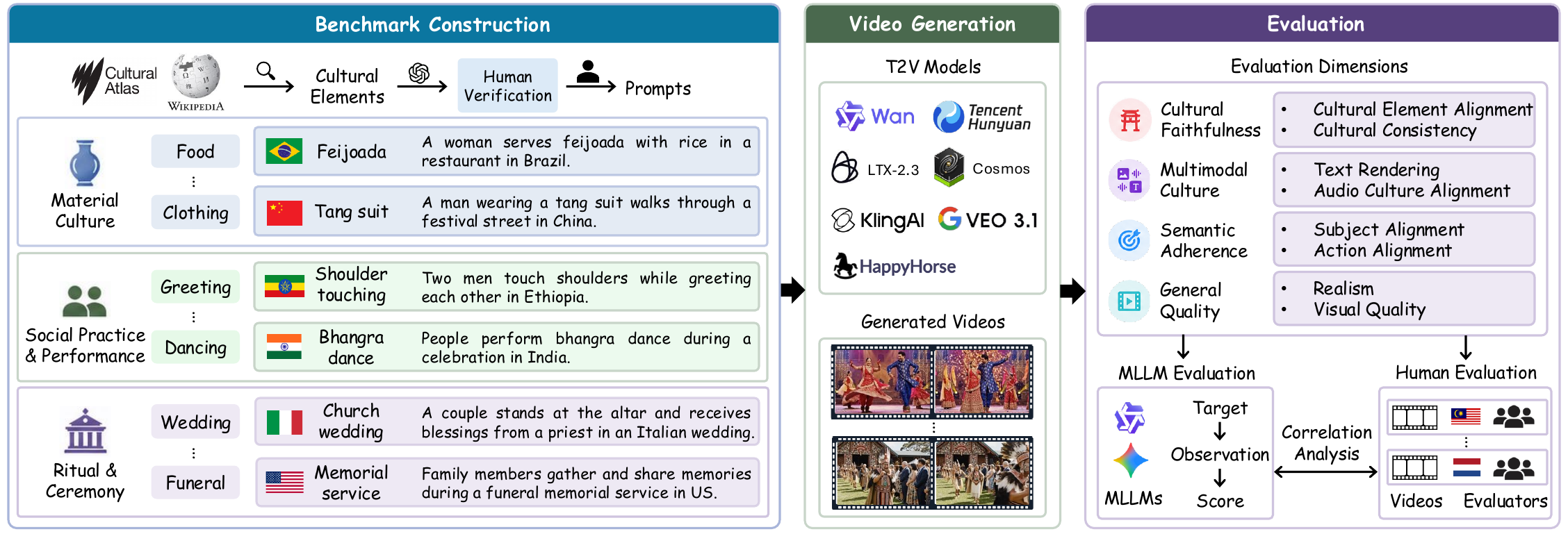}
    \vspace{-2mm}
	\caption{Framework of the CultureVidBench construction and evaluation pipeline. We collect cultural elements for each country and cultural aspect, and construct text prompts for video generation. The generated videos are evaluated by both humans and MLLMs across multiple dimensions, and their correlations are analyzed to assess the reliability of automatic evaluation.}
	\label{Fig: framework}
    \vspace{-2mm}
\end{figure*}

\section{Benchmark Construction}
The goal of CultureVidBench is to provide a comprehensive benchmark for evaluating cultural understanding in T2V generation.
This requires broad coverage of cultural regions, comprehensive representation of cultural aspects, and prompts designed for dynamic video content. In this section, we construct CultureVidBench through country selection, cultural taxonomy, and prompt generation.

\subsection{Data Source and Cultural Taxonomy}
\textbf{Country Selection}. To ensure broad cultural coverage, we select countries based on the cultural regions defined by the World Values Survey~\citep{wvs2023culturalmap}. Specifically, CultureVidBench covers 8 world cultural regions, including Confucian, West and South Asian, African-Islamic, Orthodox Europe, Catholic Europe, Protestant Europe, English-speaking, and Latin America. Based on this regional coverage, we include 12 countries across 6 continents, as shown in Figure~\ref{Fig: intro}(b). This design enables systematic evaluation of T2V models across culturally distinct societies while maintaining manageable benchmark scale.

\noindent\textbf{Cultural Taxonomy}. Culture is a multifaceted concept that includes material expressions, social practices, collective behaviors, and ritual traditions shared within a society~\citep{geertz2017interpretation}.
%To build a comprehensive benchmark, we draw on cultural items from CulturalAtlas and Wikipedia and organize the taxonomy into three major categories: material culture, social practice \& performance, and ritual \& ceremony. These categories further cover 14 cultural aspects, as shown in Figure~\ref{Fig: intro} (b).
Such dimensions are also commonly used in cultural knowledge sources such as CulturalAtlas~\citep{mosaica2024culturalatlas} and Wikipedia~\citep{wikipedia}, where cultures are described through artifacts, customs, social practices, festivals, ceremonies, and religious traditions. Based on this representation, we organize these sources into three major categories: material culture, social practice \& performance, and ritual \& ceremony. These categories further cover 14 cultural aspects, as shown in Figure~\ref{Fig: intro} (b).

\noindent\textbf{Cultural Element Collection}. For each country and cultural aspect, we systematically collect representative cultural elements from CulturalAtlas and Wikipedia. Examples of cultural elements are shown in Figure~\ref{Fig: framework}. For material culture and social practice \& performance, we mainly extract representative cultural elements. In contrast, rituals and ceremonies often involve structured social procedures that cannot be fully captured by a single element. Therefore, for the ritual \& ceremony category, we additionally extract representative sub-activities from cultural descriptions to better capture procedural cultural knowledge. For example, for the China-wedding-tea ceremony, we extract sequential activities such as kneeling before parents, serving tea, and receiving red envelopes.

\subsection{Prompt Generation}
Since CultureVidBench focuses on T2V generation, all prompts are designed to emphasize actions and interactions in dynamic scenarios. Therefore, for the collected cultural elements in each category, we construct prompts using category-specific templates. For material culture, we explicitly add human actions to embed cultural elements into video scenarios, using the template <subject> <specific action> <material cultural element> <country/cultural context>. For social practice \& performance, we focus on culturally grounded interactions, activities, and performances, and use the template <subject> <social practice \& performance> <country/cultural context>. For ritual \& ceremony, we generate prompts from representative sub-activities to better reflect culturally grounded behaviors. We use the template <subject> <sub-activity> <sub-activity> <country/cultural context>.

We provide GPT-5.4~\citep{OpenAI_GPT54_2026} with the collected cultural elements and the category-specific templates to generate three candidate prompts for each cultural element. All candidate prompts are then manually verified and revised by two human annotators guided by cultural element references to ensure linguistic naturalness, cultural consistency, diversity, and video generation feasibility.
For material culture and social practice \& performance categories, we manually select the most natural and culturally representative prompt. Since ritual and ceremony activities often contain richer procedural content, we retain two prompts with different activity focuses for each ritual element to improve diversity and coverage. In addition, to evaluate multimodal cultural rendering in T2V models, some prompts are designed with explicit visible-text or audio cues, such as festival banners and stage performance.

\subsection{Benchmark Statistics}
\begin{figure}[t]
	\centering
	\includegraphics[width=0.47\textwidth]{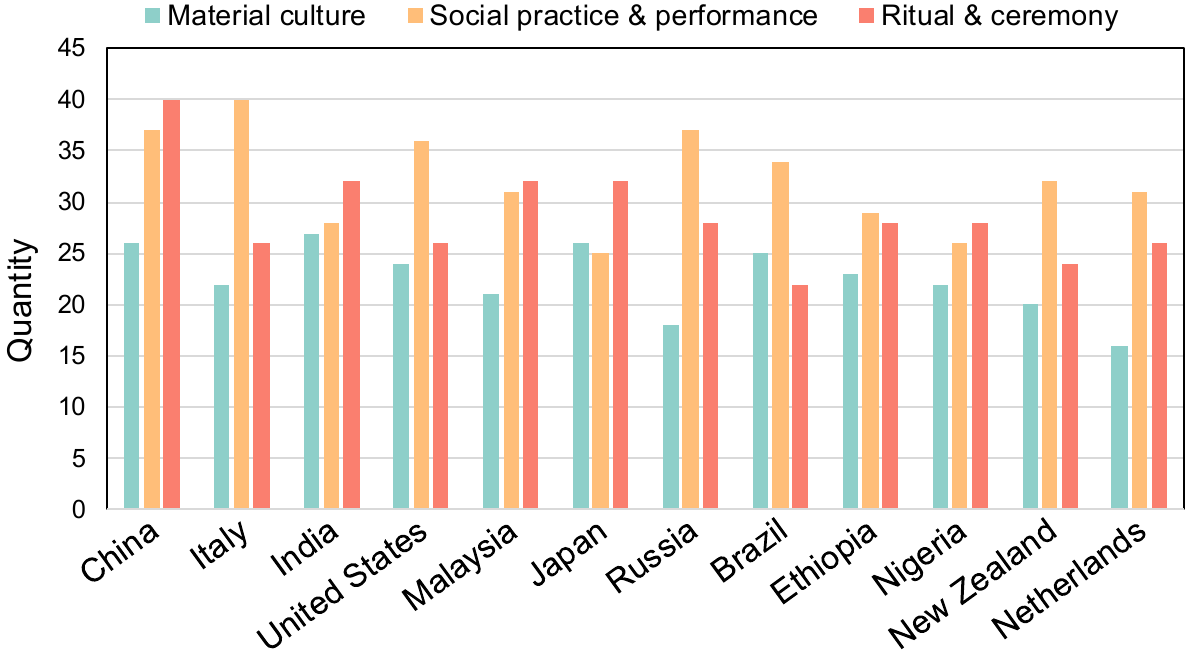}
    \vspace{-2mm}
	\caption{Prompt distribution across different countries and cultural categories.}
	\label{Fig: bar_country}
    \vspace{-3mm}
\end{figure}

CultureVidBench contains 828 cultural elements across 12 countries and 14 cultural aspects. Based on these elements, we generate and manually curate 1,000 prompts, including 272 for material culture, 386 for social practice \& performance, and 342 for ritual \& ceremony. Figure~\ref{Fig: bar_country} shows the prompt distribution across countries and cultural categories. Although the number of prompts varies slightly by country, the overall distribution remains relatively balanced. This design provides broad coverage across world cultural regions while preserving diversity in cultural categories and activities.

\section{Evaluation}
%Evaluating cultural grounding in T2V generation is challenging because cultural correctness is not equivalent to general prompt following. A generated video may be semantically aligned and visually plausible, but still misrepresent the target culture through incorrect cultural elements, mixed cultural symbols, inappropriate actions, or mismatched audio. To address these challenges, w
We design an evaluation protocol that separately assesses cultural faithfulness, multimodal cultural rendering, semantic adherence, and general quality. We further combine scalable MLLM-based automatic evaluation with human user studies, where human judgments are used to validate the reliability of automatic assessment.

\subsection{Evaluation Dimensions}
% We evaluate generated videos along 4 complementary dimensions: cultural faithfulness, multimodal cultural rendering, semantic alignment, and perceptual quality.

\noindent\textbf{Cultural Faithfulness.}
This dimension evaluates whether the generated video faithfully represents the target cultural context. It includes two aspects. (1) \textit{Cultural Element Alignment} measures whether the cultural elements specified in the prompt are correctly presented.
It evaluates not only visible cultural elements such as clothing and food, but also culturally specific interactions and procedures. For example, an Ethiopian shoulder-touching greeting should be represented with the appropriate greeting gesture rather than a generic interaction. A Japanese Shinto wedding should include key ritual elements such as the shrine setting and the correctly performed sake-sharing ceremony, rather than showing a modern wedding.
(2) \textit{Cultural consistency} measures whether the video avoids culturally inconsistent elements from other countries or cultural backgrounds. This is important because a video may include the correct cultural element while still hallucinating foreign symbols, resulting in a mixed or misleading cultural representation.

\noindent\textbf{Multimodal Cultural Rendering.}
Culture may also be expressed through visible text and audio, beyond visual objects and actions. We therefore evaluate whether culture-related multimodal content is rendered appropriately from two aspects. (1) \textit{Text Rendering} evaluates whether visible text, when present, is linguistically and culturally appropriate. (2) \textit{Audio Culture Alignment} evaluates whether culturally relevant audio, such as music, chanting, or performance sounds, matches the intended cultural background and scenario. Audio is evaluated only for aspects where cultural sound is expected, including celebration, dancing, concerts, weddings, funerals, and religious practices.

\noindent\textbf{Semantic Adherence.}
Although our benchmark focuses on cultural understanding, semantic alignment is necessary for determining whether the generated video follows the prompt. We evaluate two aspects. (1) \textit{Subject Alignment} measures whether the people or participants described in the prompt are correctly presented. (2) \textit{Action Alignment} measures whether the actions or activities described in the prompt are accurately depicted. This dimension helps distinguish cultural failures from general prompt-following failures.

\noindent\textbf{Perceptual Quality.}
We also evaluate the perceptual quality of the generated videos. This dimension includes two aspects. (1) \textit{Realism} measures whether the video appears natural and realistic, without obvious artifacts, distorted humans or objects, or unnatural motion. (2) \textit{Visual Quality} measures the overall visual presentation, including color, clarity, and aesthetic quality. These criteria help examine whether cultural failures persist even when videos are visually plausible.

\begin{table*}[t]
	\centering
	\scriptsize
	\setlength{\tabcolsep}{1pt}
	{
		\renewcommand{\arraystretch}{1.1}
		\begin{tabular}{>{\centering\arraybackslash}p{1.6cm}p{4.1cm}cccccccc}
			\hline
			\multicolumn{1}{c}{\multirow{2}{*}{\textbf{Evaluator}}}
			& \multicolumn{1}{c}{\multirow{2}{*}{\textbf{Models}}} 
			& \multicolumn{2}{c}{\textbf{Cultural Faithfulness}} 
			& \multicolumn{2}{c}{\textbf{Multimodal Cultural Rendering}} 
			& \multicolumn{2}{c}{\textbf{Semantic Adherence}} 
			& \multicolumn{2}{c}{\textbf{Perceptual Quality}} \\ 
			\cmidrule(lr){3-4} \cmidrule(lr){5-6} \cmidrule(lr){7-8} \cmidrule(lr){9-10}
			& 
			& \multicolumn{1}{c}{Alignment} 
			& \multicolumn{1}{c}{Consistency} 
			& \multicolumn{1}{c}{Text Rendering} 
			& \multicolumn{1}{c}{Audio Alignment} 
			& \multicolumn{1}{c}{Subject} 
			& \multicolumn{1}{c}{Action} 
			& \multicolumn{1}{c}{Realism} 
			& \multicolumn{1}{c}{Visual Quality} \\ 
			\hline
			
			\multirow{9}{*}{\shortstack{\textbf{Human}}}
			& \multicolumn{9}{l}{\cellcolor[HTML]{EFEFEF}\textit{Open-source models}} \\
			& LTX-2.3-22B~\citep{hacohen2026ltx} & 0.359 & 0.385 & 0.177 & 0.408 & 0.489 & 0.408 & 0.552 & 0.638\\
			& Cosmos-Predict-2.5-14B~\citep{ali2025world} & 0.403 & 0.464 & 0.178 & - & 0.545 & 0.466 & 0.409 & 0.491\\
			& HunyuanVideo-1.5~\citep{hunyuanvideo2025} & 0.475 & 0.509 & 0.102 & - & 0.584 & 0.511 & 0.550 & 0.607\\
			& Wan-2.2-14B~\citep{wan2025} & 0.485 & 0.499 & 0.170 & - & 0.579 & 0.489 & 0.560 & 0.644\\
			\cline{2-10}
			& \multicolumn{9}{l}{\cellcolor[HTML]{EFEFEF}\textit{Proprietary models}} \\
			& Kling-3.0~\citep{KlingAI2026} & 0.701 & 0.703 & 0.264 & 0.589 & 0.768 & 0.709 & 0.675 & 0.769\\
			& Veo-3.1-Fast~\citep{Veo2025} & \textbf{0.766} & 0.754 & 0.430 & \textbf{0.703} & \textbf{0.808} & 0.730 & 0.628 & 0.781\\
			& HappyHorse-1.0~\citep{happyhorse2026} & 0.742 & \textbf{0.757} & \textbf{0.434} & 0.674 & 0.802 & \textbf{0.753} & \textbf{0.698} & \textbf{0.808}\\
			\hline
			
			\multirow{9}{*}{\shortstack{\textbf{MLLM}\\Gemini-3.1-Pro}}
			& \multicolumn{9}{l}{\cellcolor[HTML]{EFEFEF}\textit{Open-source models}} \\
			& LTX-2.3-22B~\citep{hacohen2026ltx} & 0.454 & 0.581 & 0.214 & 0.566 & 0.730 & 0.502 & 0.513 & 0.697\\
			& Cosmos-Predict-2.5-14B~\citep{ali2025world} & 0.518 & 0.653 & 0.207 & - & 0.872 & 0.523 & 0.410 & 0.507\\
			& HunyuanVideo-1.5~\citep{hunyuanvideo2025} & 0.624 & 0.729 & 0.263 & - & 0.897 & 0.663 & 0.558 & 0.698\\
			& Wan-2.2-14B~\citep{wan2025} & 0.603 & 0.727 & 0.297 & - & 0.909 & 0.650 & 0.671 & 0.861\\
			\cline{2-10}
			& \multicolumn{9}{l}{\cellcolor[HTML]{EFEFEF}\textit{Proprietary models}} \\
			& Kling-3.0~\citep{KlingAI2026} & 0.883 & 0.909 & 0.402 & 0.809 & 0.992 & \textbf{0.885} & 0.776 & 0.935\\
			& Veo-3.1-Fast~\citep{Veo2025} & \textbf{0.915} & 0.931 & \textbf{0.456} & 0.783 & \textbf{0.994} & 0.874 & 0.745 & 0.926\\
			& HappyHorse-1.0~\citep{happyhorse2026} & 0.907 & \textbf{0.943} & 0.439 & \textbf{0.896} & 0.992 & 0.883 & \textbf{0.792} & \textbf{0.948}\\
			\hline
		\end{tabular}
	}
    \vspace{-1mm}
	\caption{Human evaluation and Gemini-3.1-Pro-based MLLM evaluation results. All scores are normalized to 0-1 for comparison. “-” indicates that the model does not support audio generation.}
	\label{Tab: human_auto}
    \vspace{-2mm}
\end{table*}

\subsection{MLLM Evaluation}
Text-video similarity models, such as CLIP~\citep{radford2021learning} or ViCLIP~\citep{wang2024internvid}, mainly capture coarse semantic consistency between prompts and generated videos. However, cultural evaluation requires more fine-grained reasoning over region-specific elements, contextual appropriateness, and possible foreign cultural hallucinations. Cultural cues may also appear across visual content, visible text, and audio, making multimodal reasoning necessary.

Recent MLLMs~\citep{google2026gemini31pro,OpenAI_GPT54_2026} provide a promising alternative because they can analyze visual content, interpret text in frames, and reason about audio. However, directly asking MLLMs to assign scores may lead to entangled judgments across dimensions. For example, a low-quality video may receive lower cultural scores even when the target cultural element is present. To reduce such interference, we design a target-explicit evaluation procedure that separates target identification, observation extraction, and score assignment when applicable.

For dimensions with explicit evaluation targets, including cultural element alignment, subject alignment, and action alignment, the MLLM follows three steps. (1) \textit{Target Identification}: the model identifies the expected evaluation target from the prompt, target country, metadata, and current dimension. For example, the cultural element for cultural element alignment and the expected participants for subject alignment. (2) \textit{Observation Extraction}: the model examines the generated video and extracts observations relevant to the identified target and current dimension. (3) \textit{Score Decision}: the model assigns a discrete score with a short justification based on the identified target and extracted observations.
We apply this structured protocol to all adopted MLLMs. This makes the evaluation more independent and interpretable across dimensions. Detailed prompts for MLLM-based evaluation are provided in the Appendix~\ref{Appendix: C}.

\subsection{Human Evaluation}
Human user studies provide direct judgments and serve as a reference for validating the reliability of MLLM-based evaluation. Since exhaustive human evaluation over all generated videos is time-consuming, we follow existing T2V benchmarks~\citep{feng2025tc} and evaluate a representative subset of CultureVidBench. To preserve benchmark diversity, we randomly sample one prompt under each cultural aspect from each country, resulting in 168 prompts. For each selected prompt, we evaluate videos generated by all seven adopted T2V models.

For each country, we recruit three native and culturally familiar evaluators from that country. The evaluation set for each country contains 98 generated videos from seven models. Each video is rated according to the evaluation dimensions described above using a 1-5 Likert scale. The final human score for each video is obtained by averaging the ratings from the three evaluators. Details of the human evaluation are provided in the Appendix~\ref{Appendix: C}.

\begin{table*}[t]
	\centering
	\scriptsize
	\setlength{\tabcolsep}{5pt}
	{
		\renewcommand{\arraystretch}{1.14}
		\begin{tabular}{
				p{3.3cm}
				*{16}{>{\centering\arraybackslash}p{0.33cm}}
			}
			\hline
			\multicolumn{1}{c}{\multirow{3}{*}{\textbf{Models}}} 
			& \multicolumn{4}{c}{\textbf{Cultural Faithfulness}} 
			& \multicolumn{4}{c}{\textbf{Multimodal Cultural Rendering}} 
			& \multicolumn{4}{c}{\textbf{Semantic Adherence}} 
			& \multicolumn{4}{c}{\textbf{Perceptual Quality}} \\ 
			\cmidrule(lr){2-5} \cmidrule(lr){6-9} \cmidrule(lr){10-13} \cmidrule(lr){14-17}
			& \multicolumn{2}{c}{Alignment} 
			& \multicolumn{2}{c}{Consistency} 
			& \multicolumn{2}{c}{Text Rendering} 
			& \multicolumn{2}{c}{Audio Alignment} 
			& \multicolumn{2}{c}{Subject} 
			& \multicolumn{2}{c}{Action} 
			& \multicolumn{2}{c}{Realism} 
			& \multicolumn{2}{c}{Visual Quality} \\ 
			\cline{2-17}
			& $\tau$ & $\rho$ 
			& $\tau$ & $\rho$ 
			& $\tau$ & $\rho$ 
			& $\tau$ & $\rho$ 
			& $\tau$ & $\rho$ 
			& $\tau$ & $\rho$ 
			& $\tau$ & $\rho$ 
			& $\tau$ & $\rho$ \\ 
			\hline
			ViCLIP~\citep{wang2024internvid}&0.137&0.158& - & - & - & - & - & - &0.135&0.159&0.224&0.325& - & - & - & - \\
			Qwen3-VL-32B~\citep{Qwen3-VL}&0.427&0.518&0.389&0.463&0.101&0.120& -& -& 0.375&0.446&0.396&0.470&0.188&0.223&0.154&0.178\\
			GPT-5.4 w/o Target&0.519& 0.614&0.441&0.513&0.437&0.472& -& -& 0.311& 0.363&0.395&0.467& 0.191& 0.221& 0.115&0.134 \\
			GPT-5.4~\citep{OpenAI_GPT54_2026} &\textbf{0.546}&\textbf{0.644}&0.469&0.545&0.441&0.487&-&-&0.335&0.384&0.411&0.487&0.202&0.235&0.110&0.125\\
			Gemini-3.1-Pro w/o Target &0.522& 0.616&0.516&0.607&0.513& 0.684& \textbf{0.373}&\textbf{0.441}&0.388& 0.437&0.406& 0.487&0.352& 0.426&0.353& 0.421\\
			Gemini-3.1-Pro~\citep{google2026gemini31pro} &0.529&0.629&\textbf{0.521}&\textbf{0.614}&\textbf{0.527}&\textbf{0.702}&0.364&0.431&\textbf{0.409}&\textbf{0.468}&\textbf{0.413}&\textbf{0.492}&\textbf{0.362}&\textbf{0.441}&\textbf{0.365}&\textbf{0.440}\\
			\hline
            Inter-human &0.591&0.689 &0.576&0.672 &0.606&0.694 &0.509&0.583 &0.448&0.529 &0.477&0.566 &0.449&0.535 &0.437&0.513\\ \hline
		\end{tabular}
	}
    \vspace{-2mm}
	\caption{Correlation between human and MLLM-based automatic evaluations in terms of country-averaged Kendall’s $\tau$ and Spearman’s $\rho$. The last row reports the mean inter-human correlation for reference.}
	\label{Tab: correlation}
    \vspace{-3mm}
\end{table*}

\section{Experiments}

\subsection{Experimental Setup}
We evaluate seven representative T2V models on CultureVidBench, including four open-source models, LTX-2.3-22B~\citep{hacohen2026ltx}, Cosmos-Predict-2.5-14B~\citep{ali2025world}, HunyuanVideo-1.5~\citep{hunyuanvideo2025}, and Wan-2.2-14B~\citep{wan2025}, and three proprietary models, Kling-3.0~\citep{KlingAI2026}, Veo-3.1-Fast~\citep{Veo2025}, and HappyHorse-1.0~\citep{happyhorse2026}. Cosmos-Predict-2.5-14B is originally designed for world simulation, and we use its text-to-world generation capability for evaluation. Detailed generation settings are provided in Appendix~\ref{Appendix: B}.

For automatic evaluation, we use both similarity-based and MLLM-based evaluators. ViCLIP~\cite{wang2024internvid} is used to measure coarse text-video semantic similarity, while Qwen3-VL-32B~\cite{Qwen3-VL}, GPT-5.4~\cite{OpenAI_GPT54_2026}, and Gemini-3.1-Pro~\cite{google2026gemini31pro} evaluate videos across our proposed dimensions. Qwen3-VL-32B and GPT-5.4 take 16 uniformly sampled frames as input, whereas Gemini-3.1-Pro supports direct video understanding and evaluates both visual content and audio-related cultural cues. All evaluation scores are normalized from a five-point scale to the range of 0-1 for model comparison.

\begin{figure}[t]
	\centering
	\includegraphics[width=0.48\textwidth]{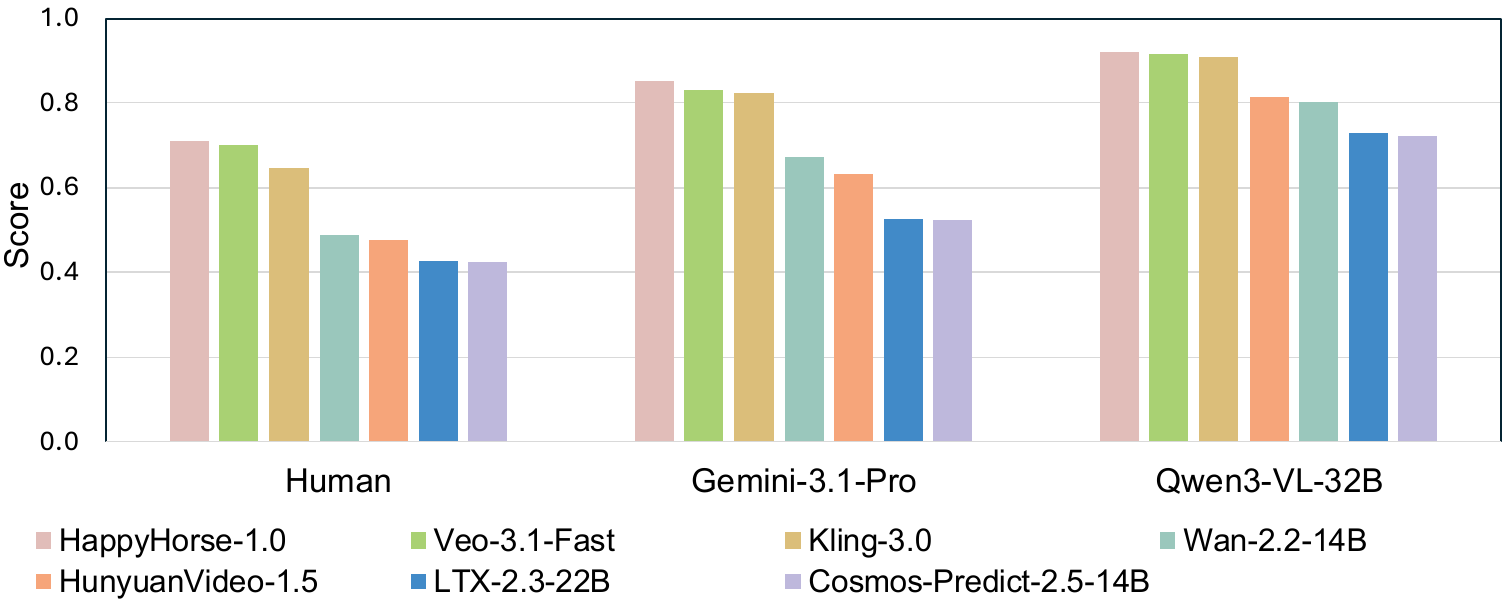}
    \vspace{-5mm}
	\caption{Overall performance rankings of T2V models from human evaluator and MLLM-based evaluators.}
	\label{Fig: bar_model}
    \vspace{-3mm}
\end{figure}

\subsection{Performance Comparison}
Table~\ref{Tab: human_auto} reports the human and MLLM-based evaluation results of seven T2V models across all dimensions. Overall, both evaluators show a clear gap between proprietary and open-source models, with proprietary models consistently achieving higher scores across most evaluation dimensions.
Moreover, current T2V models still show limited cultural understanding. Compared with semantic adherence and perceptual quality, the scores on cultural faithfulness and multimodal cultural rendering are generally lower, indicating that models can generate visually plausible and semantically related videos while missing culturally specific details. Multimodal cultural rendering is particularly challenging, as even the strongest proprietary model, HappyHorse-1.0, still shows clear limitations in producing culturally appropriate visible text.

To examine whether MLLM-based evaluators are consistent with human judgments, Table~\ref{Tab: correlation} reports the correlations between automatic and human evaluation using country-averaged Kendall's $\tau$ and Spearman's $\rho$, with inter-human agreement included as a reference.
Compared with ViCLIP, MLLM-based evaluators show higher correlations with human evaluation, suggesting the need for fine-grained multimodal reasoning in cultural evaluation. Moreover, our target-explicit structured evaluation further improves both Gemini-3.1-Pro and GPT-5.4 on dimensions with clearly defined evaluation targets, such as cultural element alignment, subject alignment, and action alignment.

\begin{figure}[t]
	\centering
	\includegraphics[width=0.46\textwidth]{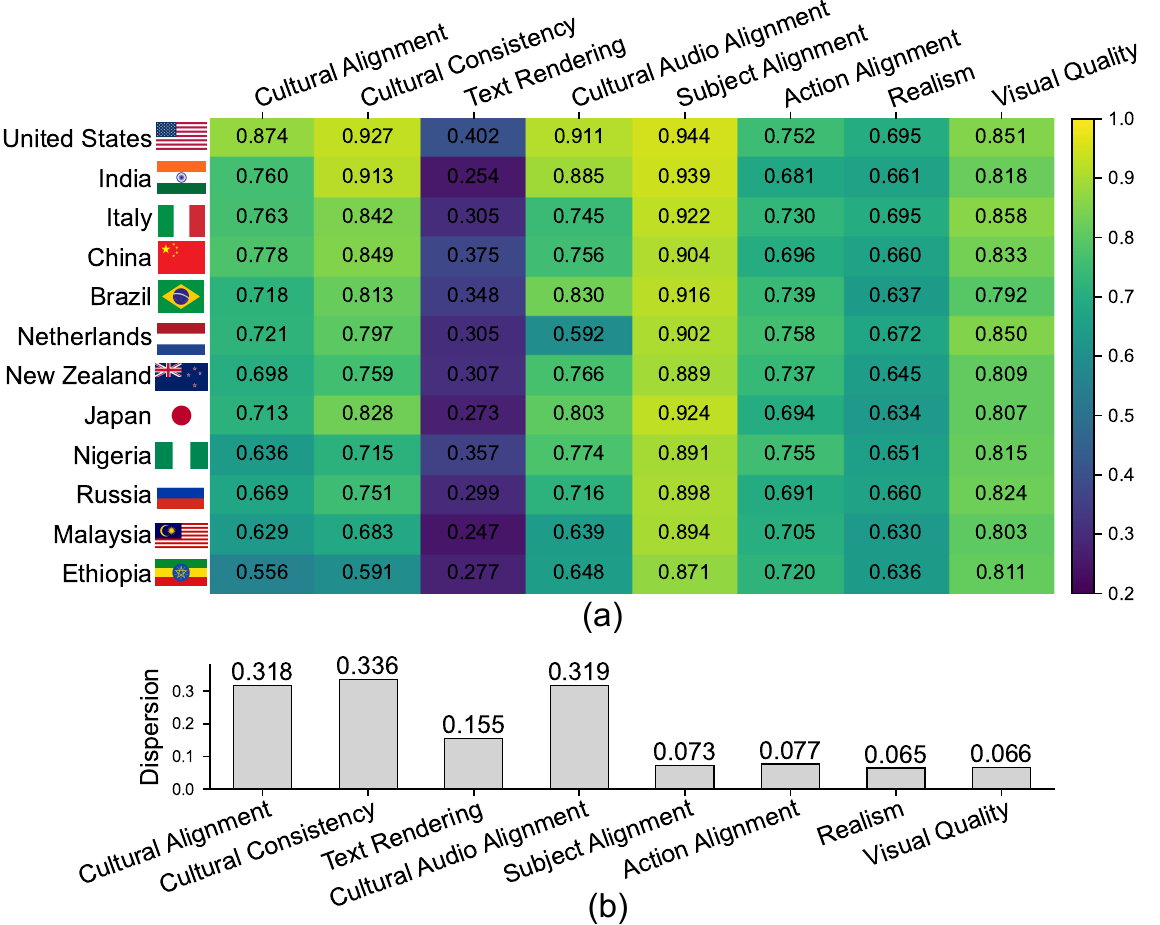}
    \vspace{-2mm}
	\caption{(a) Country-level results averaged over all T2V models evaluated by Gemini-3.1-Pro. (b) Dispersion denotes the gap between the highest and lowest country-level scores.}
	\label{Fig: heatmap}
    \vspace{-3mm}
\end{figure}

Among automatic evaluators, Gemini-3.1-Pro with full-video input achieves the highest overall correlation with human judgments, showing the benefit of direct video-level input for multimodal understanding and cultural reasoning. However, its correlations remain relatively limited on more subjective dimensions, such as realism and visual quality, suggesting that these aspects are still challenging for MLLM-based evaluators. Notably, inter-human agreement is also limited on these subjective dimensions, partly because the fine-grained rating scale requires annotators to distinguish subtle differences between adjacent scores.

We further visualize the model rankings produced by human evaluation, Gemini-3.1-Pro, and Qwen3-VL-32B across all evaluation dimensions in Figure~\ref{Fig: bar_model}. Gemini-3.1-Pro produces rankings consistent with human judgments, while Qwen3-VL-32B can also distinguish proprietary from open-source models. These results demonstrate that MLLM-based evaluation can provide a reliable and scalable approximation of human assessment.

\begin{figure}[t]
	\centering
	\includegraphics[width=0.47\textwidth]{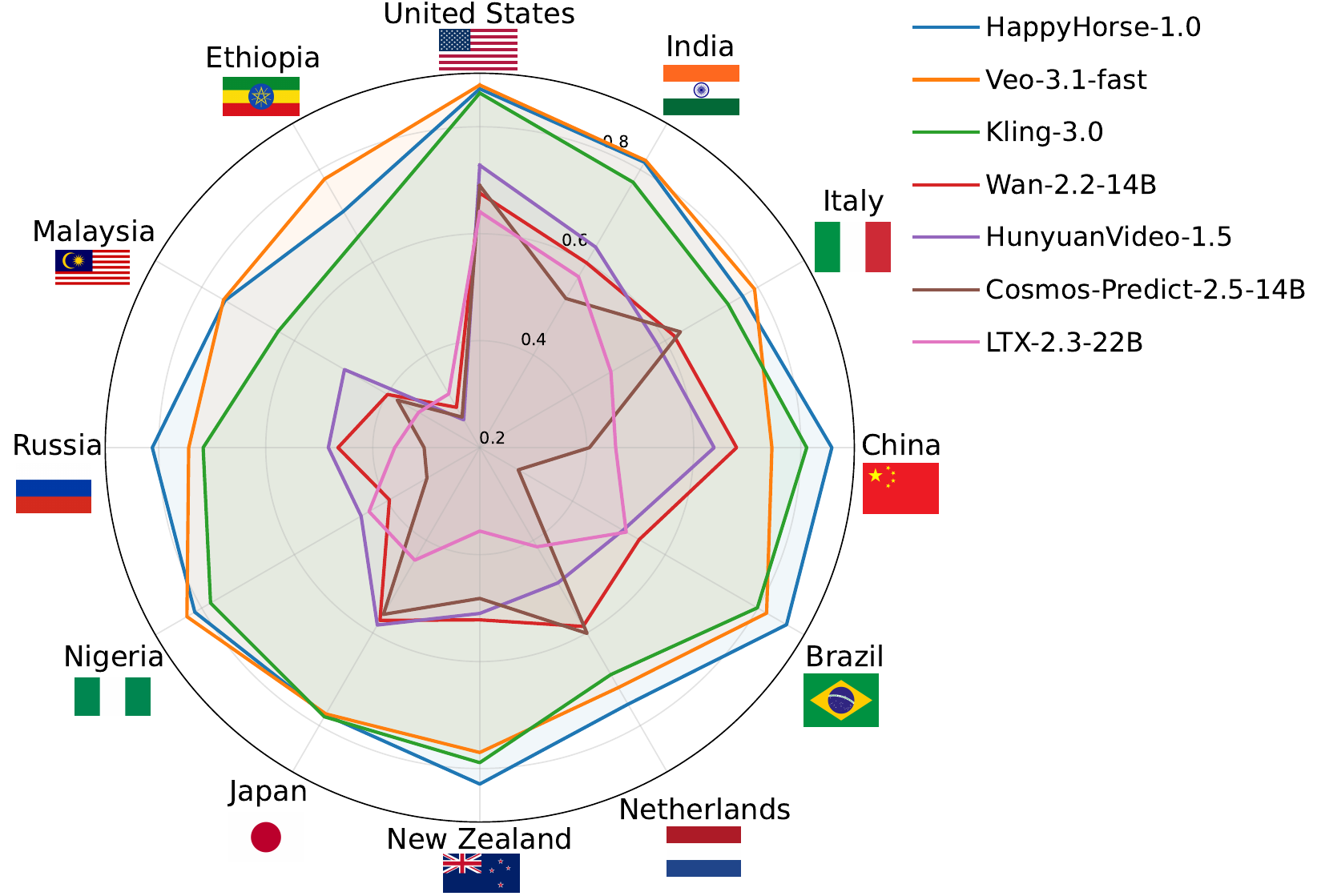}
    \vspace{-1mm}
	\caption{Country-level cultural scores of different T2V models evaluated by Gemini-3.1-Pro. Scores are averaged over cultural faithfulness and multimodal cultural rendering.}
	\label{Fig: country_poly}
    \vspace{-2mm}
\end{figure}

\subsection{Cross-country Cultural Performance}
We then analyze the performance of T2V models across countries. Figure~\ref{Fig: heatmap} (a) shows the country-level results averaged over all evaluated models across different evaluation dimensions. To better quantify cross-country variation, we also report the dispersion of each evaluation dimension in Figure~\ref{Fig: heatmap} (b). The results show that current T2V models perform relatively consistently across countries on semantic adherence and perceptual quality. In contrast, culture-related dimensions exhibit much larger cross-country disparities. We can observe that T2V models tend to achieve stronger performance on high-resource cultural contexts, such as the United States, while performing worse on relatively underrepresented cultural contexts, such as Ethiopia and Malaysia.

We further examine the country-level cultural scores of different T2V models in Figure~\ref{Fig: country_poly}. The results reveal clear cross-country variation across models. Open-source models show larger performance fluctuations and generally perform worse on relatively underrepresented cultural contexts, including Ethiopia, Malaysia, and Nigeria. Proprietary models achieve higher and more stable cultural scores overall, but they still show visible drops on these countries.

\begin{figure}[t]
	\centering
	\includegraphics[width=0.47\textwidth]{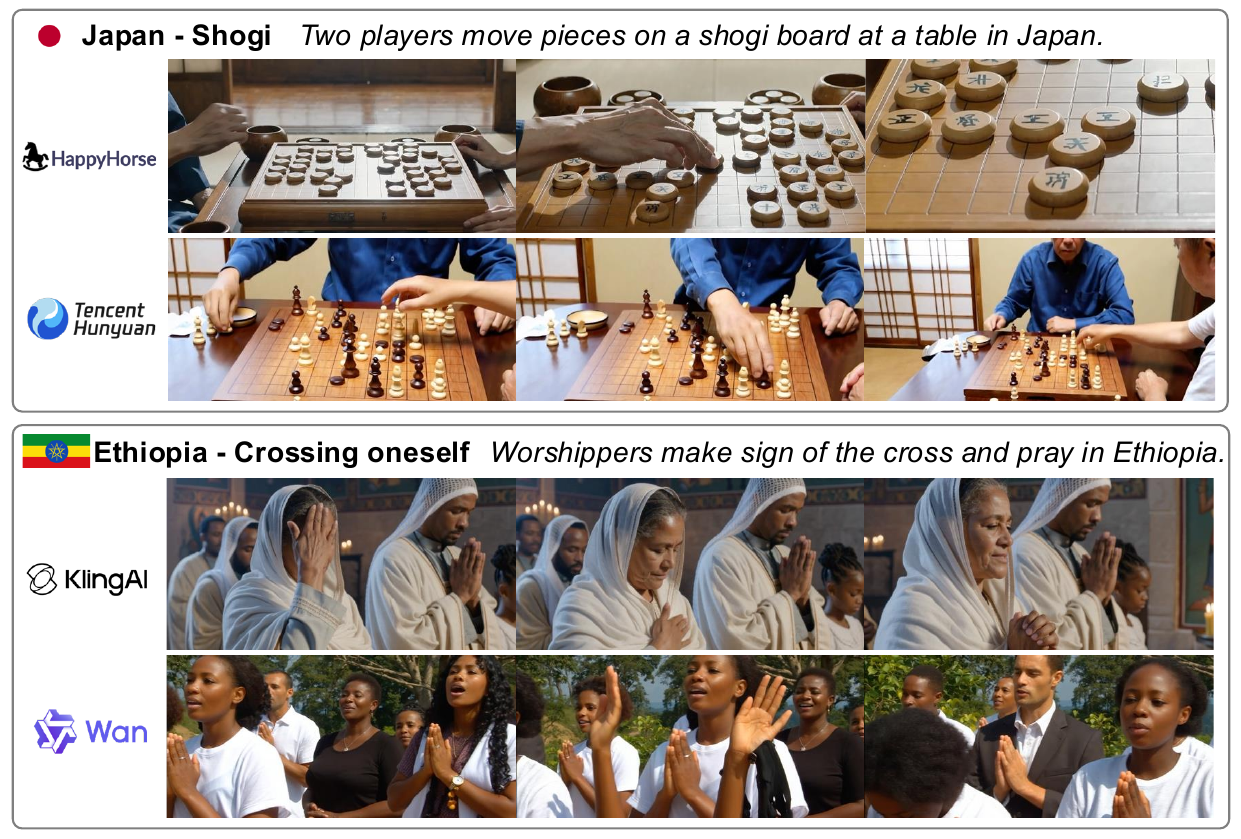}
    \vspace{-1mm}
	\caption{Cultural errors in videos generated by T2V models, such as incorrect visible text and inaccurate cultural elements.}
	\label{Fig: case}
    \vspace{-2mm}
\end{figure}

To provide a more intuitive analysis, Figure~\ref{Fig: case} shows representative cultural errors made by different T2V models. In the Japan-Shogi example, HappyHorse-1.0 captures the shogi scenario, but still produces inaccurate visible text on the pieces, while HunyuanVideo-1.5 generates a chess-like scene instead of the correct Shogi element. In the Ethiopia-Crossing oneself example, all models struggle to faithfully represent the culturally specific religious gesture, which requires pinching the fingers together when making the sign of the cross~\citep{wikipedia}. Overall, these findings show that current T2V models can often capture the coarse semantic content of the prompt, such as ``two players'' and ``move pieces'', but still have limitations in culturally grounded generation, especially for underrepresented cultural contexts.

\subsection{Cultural Category Analysis}
\begin{figure}[t]
	\centering
	\includegraphics[width=0.47\textwidth]{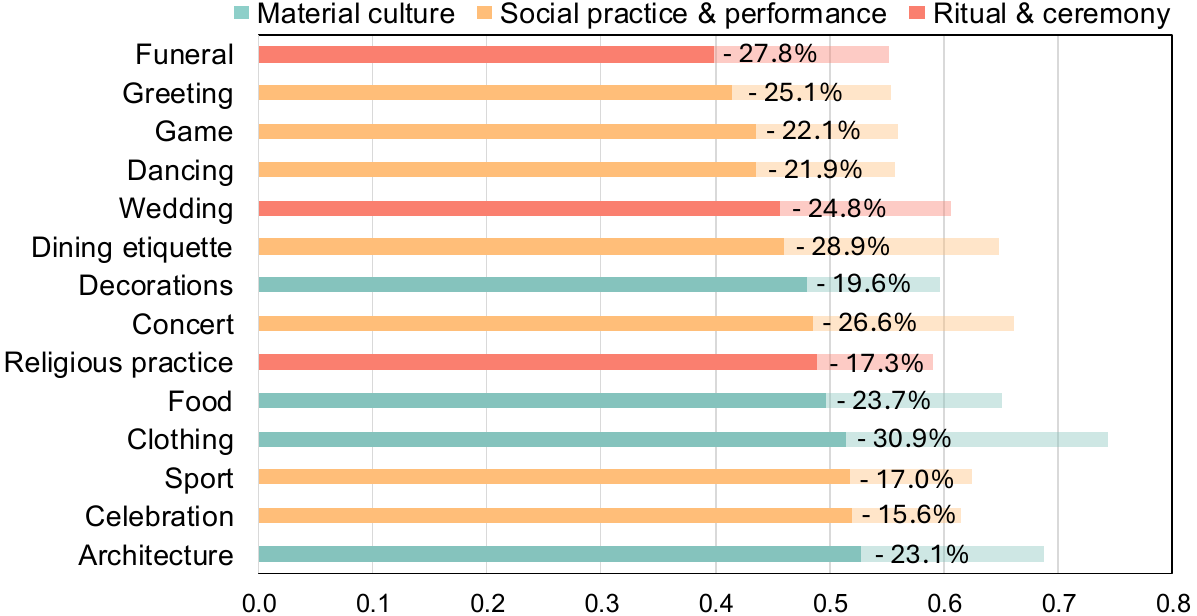}
    \vspace{-1mm}
    \caption{Cultural scores (dark-colored bars) and semantic adherence scores (light-colored bars) across cultural aspects evaluated by human. Cultural scores are averaged over cultural faithfulness and multimodal cultural rendering. Percentages indicate the relative decrease from semantic adherence to cultural score.}
	\label{Fig: category}
    \vspace{-2mm}
\end{figure}

We further analyze model performance across different cultural categories and aspects in Figure~\ref{Fig: category}. Overall, semantic adherence is consistently higher than cultural performance across all cultural aspects, showing the limitations of current T2V models in culturally grounded generation.
Among the three cultural categories, material culture generally achieves stronger cultural performance, especially for aspects such as architecture, clothing, and food, which are strongly associated with distinctive visual appearances. In contrast, social practice \& performance and ritual \& ceremony aspects are more challenging, as they require models to capture culturally grounded actions, interactions, and procedures rather than only visual objects. This suggests that cultural video generation remains particularly difficult when culture is expressed through human activities and ritual processes.

\section{Conclusion}
In this paper, we presented CultureVidBench, a comprehensive benchmark for evaluating cultural understanding in T2V generation. CultureVidBench covers diverse countries, cultural regions, and cultural aspects, and evaluates generated videos across cultural faithfulness, multimodal cultural rendering, semantic adherence, and perceptual quality through both human evaluation and MLLM-based assessment.
We evaluate both proprietary and open-source T2V models and validate the reliability of MLLM-based evaluation. Our results show that current T2V models achieve strong semantic adherence and visual quality, but still struggle with cultural faithfulness and multimodal cultural rendering, especially for underrepresented regions, rituals, and multimodal cultural cues such as visible text and audio.
These findings highlight the importance of systematic cultural evaluation, and CultureVidBench can facilitate future research toward more culturally aware and trustworthy T2V generation systems.

\section*{Limitations}
CultureVidBench currently covers 12 countries selected to span multiple continents and cultural regions, but it does not exhaustively cover all countries or cultural contexts. This design allows us to evaluate cultural understanding across diverse and representative scenarios, but may still miss many local, regional, and minority cultural practices. In future work, CultureVidBench can be further expanded to include a larger set of countries, more fine-grained cultural contexts, and broader community or expert contributions.

\section*{Ethical Considerations}
CultureVidBench is designed for evaluating cultural understanding in T2V generation. During benchmark construction, we conducted safety checks to avoid prompts involving harmful, offensive, violent, or discriminatory content. CultureVidBench will be released for research and evaluation purposes only. The selected prompts are intended to evaluate whether models can faithfully represent culturally grounded objects, actions, rituals, visible text, and audio cues, rather than to rank cultures themselves.

For human evaluation, we use a representative subset of CultureVidBench to reduce the annotation burden while maintaining coverage across countries and cultural aspects. Human evaluators are recruited through Prolific and are compensated for their participation. The evaluation focuses on generated videos rather than personal information, and no private or sensitive participant data is collected beyond what is required by the study platform. Participants were informed that their responses would be used for research purposes to evaluate generated videos.

\section*{Acknowledgments}
This research/project is supported by the Ministry of Education (MOE), Singapore, under its Academic Research Fund (AcRF) Tier 2 (Proposal ID: T2EP20125-0048). Any opinions, findings and conclusions or recommendations expressed in this material are those of the authors and do not reflect the views of the Ministry of Education, Singapore. This research/project is also supported by the National Research Foundation Singapore under the AI Singapore Programme (AISG Award No: AISG3-RPGV-2025-016).

\bibliography{custom}

\clearpage
\appendix
\section{CultureVidBench Details}
\label{Appendix: A}
CultureVidBench contains 1,000 prompts across 14 cultural aspects. To reduce cultural stereotyping, we avoid prompts based on physical appearance, racialized descriptions, or essentialized portrayals of people. Instead, we prioritize concrete cultural objects, practices, performances, and ritual activities grounded in cultural sources. The prompts are concise yet descriptive, with an average length of 12.8 words, providing sufficient cultural context for video generation while avoiding unnecessary linguistic complexity.

To further position CultureVidBench among existing cultural benchmarks, Table~\ref{Tab: benchmarks} compares it with representative benchmarks for cultural evaluation. Existing benchmarks mainly focus on text-to-image generation, covering cultural artifacts, social activities, multilingual visual concepts, or cultural stereotypes. While these benchmarks provide valuable resources for evaluating static cultural representation, they do not address the dynamic and multimodal nature of video generation.
In contrast, CultureVidBench is designed specifically for T2V generation. Unlike prior benchmarks that mainly focus on static visual content or limited cultural dimensions, CultureVidBench emphasizes culturally grounded actions, social interactions, ritual procedures, visible text, and audio cues in video generation.

Table~\ref{Tab: benchmark_prompt_comparison} further compares the prompt design of CultureVidBench with CUBE-1K~\citep{kannen2024beyond} and CulturalFrames~\cite{nayak2025culturalframes}. While prior benchmarks mainly use static descriptions or simple human actions, CultureVidBench includes culturally grounded interactions, multi-step ritual activities, and explicit visible-text cues. Such prompt designs enable a more comprehensive evaluation of whether T2V models can generate culturally appropriate human actions and social interactions, follow the temporal and procedural structure of cultural activities, and accurately render multimodal cultural elements in both visual and audio modalities.

\begin{table}[t]
	\centering
	\scriptsize
    \setlength{\tabcolsep}{4.5pt}
	{
		\renewcommand{\arraystretch}{1.18}
		\begin{tabular}{lcccc}
			\hline
			\textbf{Models}   & \textbf{Resolution}& \textbf{Frames} & \textbf{Audio} & \textbf{Duration (s)} \\ \hline
			LTX-2.3-22B& 1920$\times$1088   & 121    & \checkmark  & 5        \\
			Cosmos-Predict-2.5-14B& 1280$\times$704  & 93    &   & 6       \\
			HunyuanVideo-1.5 & 1280$\times$720  & 121    & & 5       \\
			Wan-2.2-14B & 1280$\times$720  & 81     & & 5      \\ \hline
			Kling-3.0& 1280$\times$720 & 121    &\checkmark& 5      \\
			Veo-3.1-Fast& 1280$\times$720  & 144 &\checkmark & 6    \\ 
			Happyhorse-1.0& 1280$\times$720  & 121    &\checkmark & 5    \\ \hline
		\end{tabular}
	}
	\caption{Generation settings of the evaluated T2V models, including resolution, total number of frames, audio availability, and video duration.}
	\label{Tab: generation}
\end{table}

\begin{table*}[t]
	\centering
	\scriptsize
	\setlength{\tabcolsep}{3pt}
	\renewcommand{\arraystretch}{1.2}
	\resizebox{\textwidth}{!}{
		\begin{tabular}{lccccccccc}
			\hline
			\textbf{Benchmark}
			& \textbf{Model}
			& \textbf{\# Countries}
			& \textbf{\# Cultural Aspects}
			& \textbf{\# Prompts}
			& \makecell{\textbf{Cultural}\\\textbf{Faithfulness}}
            & \makecell{\textbf{Cultural}\\\textbf{Stereotype}}
			& \makecell{\textbf{Perceptual}\\\textbf{Quality}}
			& \makecell{\textbf{Semantic}\\\textbf{Alignment}}
			& \makecell{\textbf{Multimodal}\\\textbf{Rendering}}\\
			\hline
            ViSAGe~\citep{jha2024visage} & T2I&135&1&405& & \checkmark & & & \\
			CUBE-1K~\citep{kannen2024beyond} & T2I& 8& 3& 1,000& \checkmark&  & \checkmark &  & \\
			CultureBench~\citep{shi2025culture} & T2I & 15&-& 7,932& \checkmark & &\checkmark&\checkmark& \\
            CulturalFrames~\citep{nayak2025culturalframes}&T2I &10 &5 &983 &\checkmark &\checkmark &\checkmark &\checkmark & \\
            CultDiff~\citep{bayramli2025diffusion}&T2I &10 &3 &1,500 &\checkmark &&\checkmark &\checkmark & \\
            CuRe~\cite{rege2025cure}&T2I &64 &32 &300 &\checkmark & &\checkmark &\checkmark & \\
            CULTIVate~\citep{malakouti2026culture} & T2I & 16& 9& 576 & \checkmark &  &  &\checkmark & \\
			CultureVidBench (ours)& T2V & 12& 14& 1,000& \checkmark & &\checkmark & \checkmark & \checkmark \\
			\hline
		\end{tabular}
	}
	\caption{Comparison of CultureVidBench with existing cultural benchmarks.}
	\label{Tab: benchmarks}
\end{table*}

\begin{table*}[t]
	\centering
	\scriptsize
	\setlength{\tabcolsep}{5pt}
	\renewcommand{\arraystretch}{2}
	\resizebox{\textwidth}{!}{
		\begin{tabular}{llll}
			\hline
			&\textbf{CUBE-1K} &\textbf{CulturalFrames} &\textbf{CultureVidBench (Ours)}
			\\
			\hline
			Evaluated Model
			& T2I
			& T2I
			& T2V
			\\
			\makecell[l]{Prompt with Human Actions}
			& $\times$
			& \makecell[l]{A bride and groom exchange vows\\
				at a Hindu wedding in India.}
			& \makecell[l]{Two players move pieces on a shogi board at a table\\ in Japan.}
			\\

			\makecell[l]{Prompt with Multi-Step Activities}
			& $\times$
			& $\times$
			& \makecell[l]{A bride and groom sit near a sacred fire and offer ritual\\ items into the flames during a wedding in India.}
			\\

			\makecell[l]{Prompt with Text Cues}
			& $\times$
			& $\times$
			& \makecell[l]{Families decorate homes for Christmas with a\\ festive banner in the United States.}
			\\
            \makecell[l]{Prompt with Audio Cues}
			& $\times$
			& $\times$
			& \makecell[l]{Performers sing and dance to Afrobeats music\\ on stage in Nigeria.}
			\\
			\hline
		\end{tabular}
	}
	\caption{Prompt comparison of CultureVidBench with existing T2I cultural benchmarks.}
	\label{Tab: benchmark_prompt_comparison}
\end{table*}

\section{T2V Generation Setting}
\label{Appendix: B}
For each T2V model, we generate videos for all prompts in CultureVidBench, resulting in 7,000 videos in total. We follow the official implementations and default generation settings of each model whenever possible. Table~\ref{Tab: generation} reports the detailed settings, including resolution, number of frames, audio availability, and video duration. For LTX-2.3-22B, we use its recommended two-stage generation pipeline. For Cosmos-Predict-2.5-14B, which is designed for world simulation, we use its text-to-world generation function.

\section{Evaluation}
\label{Appendix: C}
\subsection{Human Evaluator Selection}
We construct the human evaluation questionnaires using Qualtrics\footnote{\url{https://www.qualtrics.com/}} and recruit evaluators through Prolific\footnote{\url{https://www.prolific.com/}}. Participants are asked to evaluate videos corresponding to their own cultural background. To improve cultural reliability and reduce language barriers, each questionnaire is translated into the official language of the corresponding country. We further screen participants based on nationality, current residence, and primary language to increase the likelihood that they are familiar with the target cultural context. To ensure annotation quality, we include attention-check questions and remove invalid submissions that fail these checks. Participants are compensated at an hourly rate of \pounds6.

\begin{figure*}[t]
	\centering
	\includegraphics[width=0.96\textwidth]{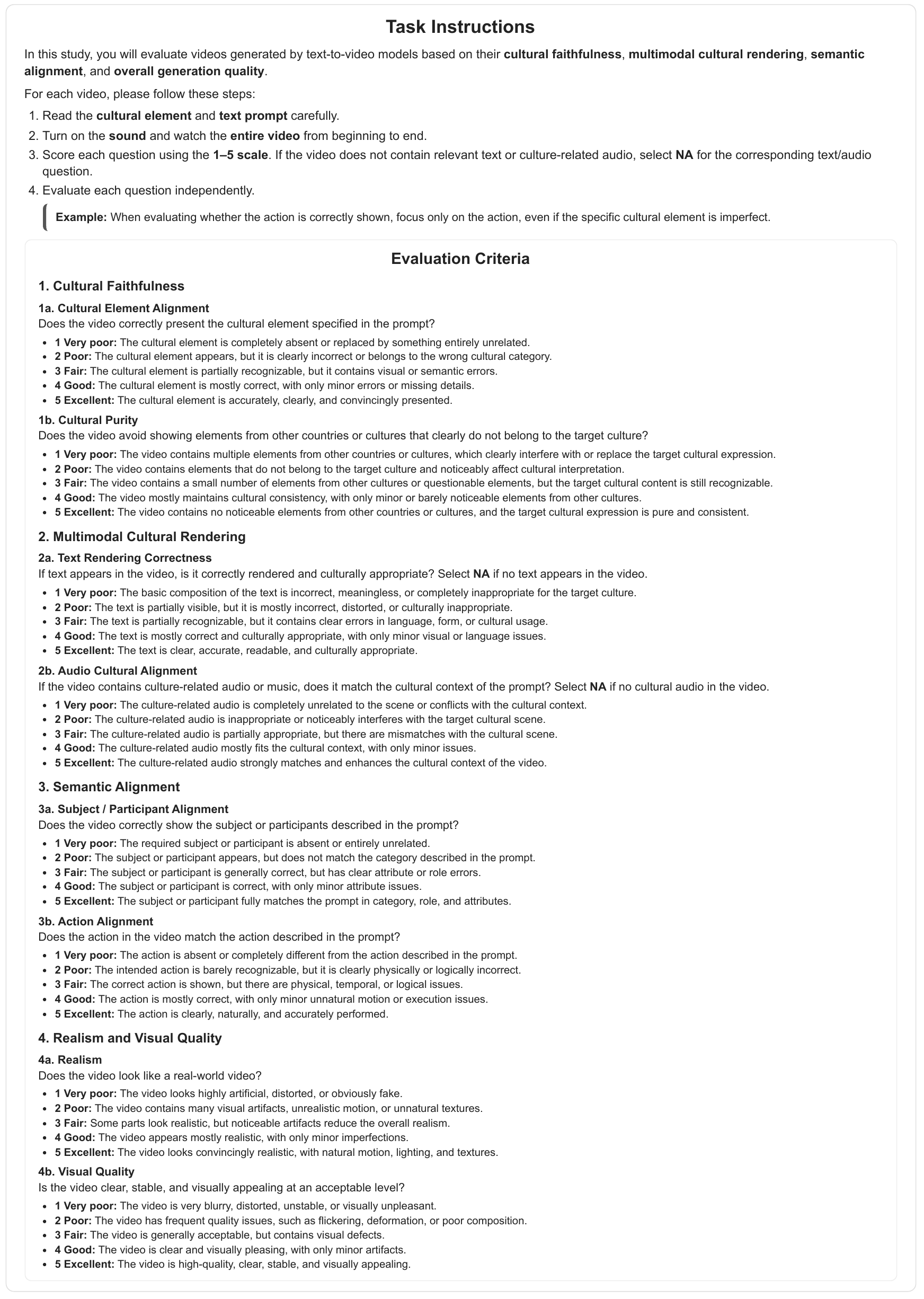}
	\caption{Task instructions and evaluation criteria in the human evaluation interface.}
	\label{Fig: criteria}
\end{figure*}

\begin{figure*}[t]
	\centering
	\includegraphics[width=0.96\textwidth]{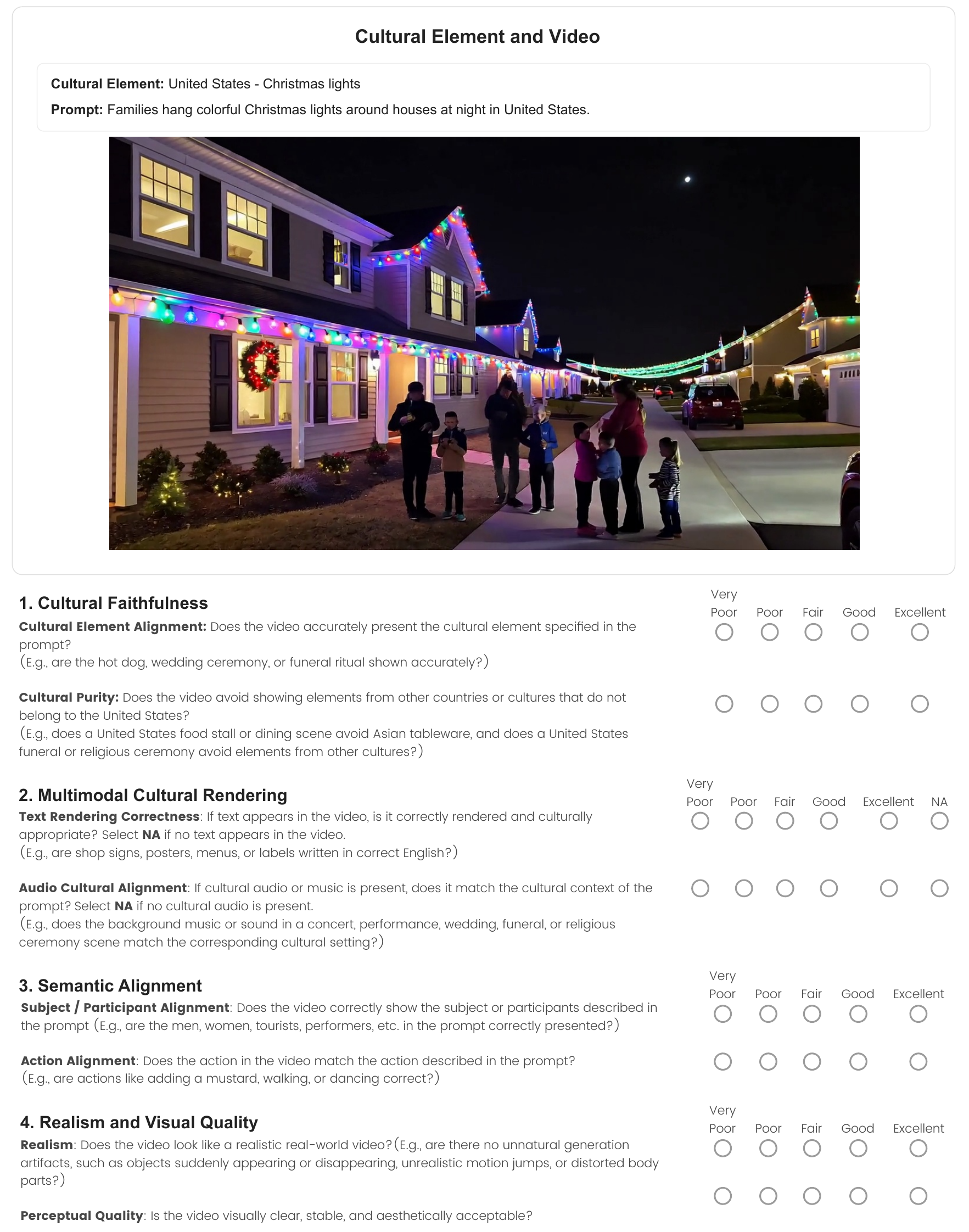}
	\caption{Human evaluation interface used for rating generated videos.}
	\label{Fig: interface}
\end{figure*}

Figure~\ref{Fig: criteria} shows the task instructions and detailed evaluation criteria provided to participants during the human evaluation. Figure~\ref{Fig: interface} shows the rating interface, where participants view the cultural element, prompt, and generated video, and then assign scores across the evaluation dimensions.

\subsection{MLLM-based Evaluation Details}
We conduct MLLM-based evaluation to provide scalable automatic assessment for generated videos. We adopt both open-source and proprietary MLLMs, including Qwen3-VL-32B, GPT-5.4, and Gemini-3.1-Pro. Among them, Gemini-3.1-Pro supports full-video input and audio understanding, which allows it to directly evaluate both visual and audio-related cultural dimensions. The detailed prompts used for Gemini-3.1-Pro-based evaluation are provided in Table~\ref{Tab: prompt}.

\begin{figure*}[t]
	\centering
	\includegraphics[width=0.96\textwidth]{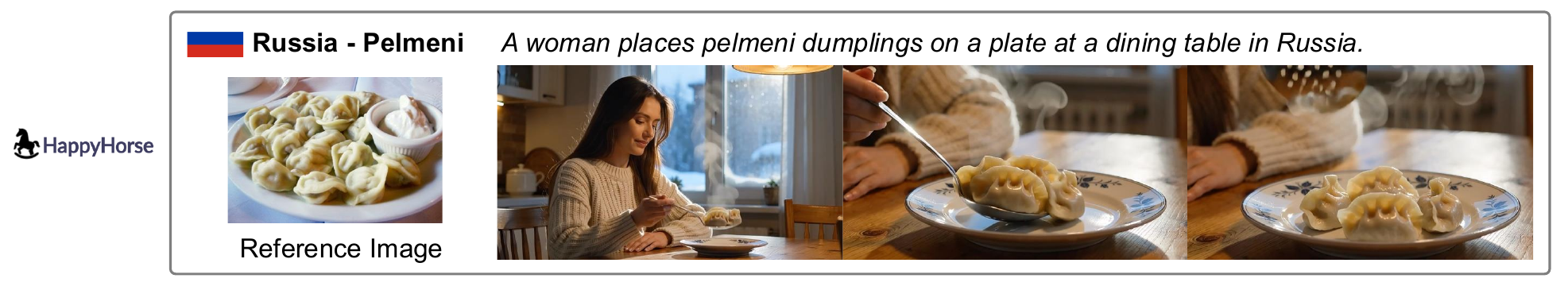}
	\vspace{3mm}
	\scriptsize
	\setlength{\tabcolsep}{4pt}
	\renewcommand{\arraystretch}{1.25}
	\begin{tabular}{p{1.6cm}p{3.2cm}p{3.2cm}p{3.0cm}p{3.0cm}}
		\hline
		& \multicolumn{1}{c}{\textbf{Cultural Element Alignment}} 
		& \multicolumn{1}{c}{\textbf{Cultural Consistency}} 
		& \multicolumn{1}{c}{\textbf{Subject Alignment}} 
		& \multicolumn{1}{c}{\textbf{Action Alignment}} \\
		\hline
		Target
		& The video should show pelmeni, traditional Russian meat dumplings. 
		& -- 
		& The video should feature a woman. 
		& The woman should be placing pelmeni dumplings on a plate. \\
		Observation
		& The dumplings appear large and crescent-shaped with crimped edges, rather than the small, round, ear-like shape of pelmeni. 
		& The dumplings do not appear to belong to the target local culture, while other elements, such as the dining table and snowy winter setting, are consistent with the scene. 
		& A woman is clearly visible sitting at the table. 
		& The woman is placing the dumplings with a spoon. \\
		MLLM Score
		& 2-Poor 
		& 3-Fair 
		& 5-Excellent 
		& 5-Excellent \\
		\hline
        Human Score
		& 2.3 
		& 3.0 
		& 5.0 
		& 5.0 \\
        \hline
	\end{tabular}
	\caption{Example of target-explicit evaluation using Gemini-3.1-Pro. The evaluator identifies the target requirement, observes the generated video, and assigns a score for each dimension. Human scores are averaged from three evaluators. We also provide a Wikipedia reference image for the target cultural element.}
	\label{Fig: eval_example}
\end{figure*}

To make the evaluation more interpretable and reduce interference across different dimensions, we use a target-explicit structured evaluation protocol. This structure encourages the MLLM to evaluate dimensions independently rather than relying on a general impression of the video.
Figure~\ref{Fig: eval_example} presents an example of target-explicit evaluation using Gemini-3.1-Pro. In this example, the cultural element should be pelmeni, a traditional Russian meat dumpling, as shown in the reference image. The MLLM correctly observes that the generated dumplings are large and crescent-shaped with crimped edges, which do not match the small, round, ear-like shape of pelmeni, and therefore assigns a low score for cultural element alignment. At the same time, it gives high scores for subject alignment and action alignment, since the woman is clearly visible and is placing dumplings on a plate. The MLLM scores are close to the averaged human scores from three evaluators, showing that the structured protocol helps MLLMs provide dimension-specific judgments that are consistent with human assessment.

\subsection{Correlation Analysis}
We measure the consistency between MLLM-based evaluation and human evaluation using Kendall's $\tau$ and Spearman's $\rho$. These rank-based metrics indicate whether automatic evaluators produce model rankings that agree with human judgments.
Since human evaluation is conducted separately for each country, directly computing a single global correlation may be affected by cross-country differences in evaluator scoring styles. Therefore, we compute correlations independently within each country and then average the country-level correlations:
\begin{equation}
	\tau = \frac{1}{|\mathcal{C}|} \sum\nolimits_{c \in \mathcal{C}} \tau_c,
	\quad
	\rho = \frac{1}{|\mathcal{C}|} \sum\nolimits_{c \in \mathcal{C}} \rho_c,
\end{equation}
where $\mathcal{C}$ denotes the set of evaluated countries. $\tau_c$ and $\rho_c$ denote Kendall's $\tau$ and Spearman's $\rho$ computed within country $c$, respectively.

\begin{table*}[t]
	\centering
	\scriptsize
	\begin{tabular}{p{0.96\linewidth}}
		\toprule
		You are an expert evaluator for cultural capability in text-to-video generation. You are given a 5-second AI-generated video.\\
        \\
        Your task is to evaluate how well the video matches the prompt and the target cultural context. \\
		\\
		Metadata:\\
        - Target country/culture: Netherlands\\
        - Cultural element: Cheese\\
        - Video prompt: A vendor cuts large cheese wheels into slices at an open-air market with a sign board in the Netherlands.\\
		\\
        Rules:\\
        - Evaluate only what is visible in the video.\\
        - Use the full 1-5 scale.\\
        - For text rendering, use "NA" if no visible text appears.\\
        - For audio cultural alignment, use "NA" only if the video has no audible audio, or if the audio is purely non-cultural ambient sound.\\
        - Evaluate each question independently.\\
        - The requirement fields should describe the evaluation target, not the model's visual observation.\\
        - The observation fields should describe only visible evidence from the frames.\\
        \\
		Evaluation criteria:\\
		\{Criteria\}\\
		\\
		Output Format:\\
        Return strictly valid JSON only. Do not include markdown, explanations outside JSON, or extra text.\\
		\{\\
		"1a": \{
            "requirement": "Expected cultural element.",
            "observation": "",
            "justification": "",
            "score": 1,
        \},\\
        "1b": \{
            "observation": "",
            "justification": "",
            "score": 1,
        \},\\
        "2a": \{
            "observation": "",
            "justification": "",
            "score": 1,
        \},\\
        "2b": \{
            "audio observation": "",
            "justification": "",
            "score": 1,
        \},\\
        "3a": \{
            "requirement": "Expected subject or participants.",
            "observation": "",
            "justification": "",
            "score": 1,
        \},\\
        "3b": \{
            "requirement": "Expected action.",
            "observation": "",
            "justification": "",
            "score": 1,
        \},\\
        "4a": \{
            "observation": "",
            "justification": "",
            "score": 1,
        \},\\
        "4b": \{
            "observation": "",
            "justification": "",
            "score": 1,
        \}\\
		\}
		\\
		\bottomrule
	\end{tabular}
	\caption{Prompt template for target-explicit MLLM-based evaluation. Criteria are the same as those used in the human-evaluation interface.}
	\label{Tab: prompt}
\end{table*}

\begin{figure*}[h]
	\centering
	\includegraphics[width=0.98\textwidth]{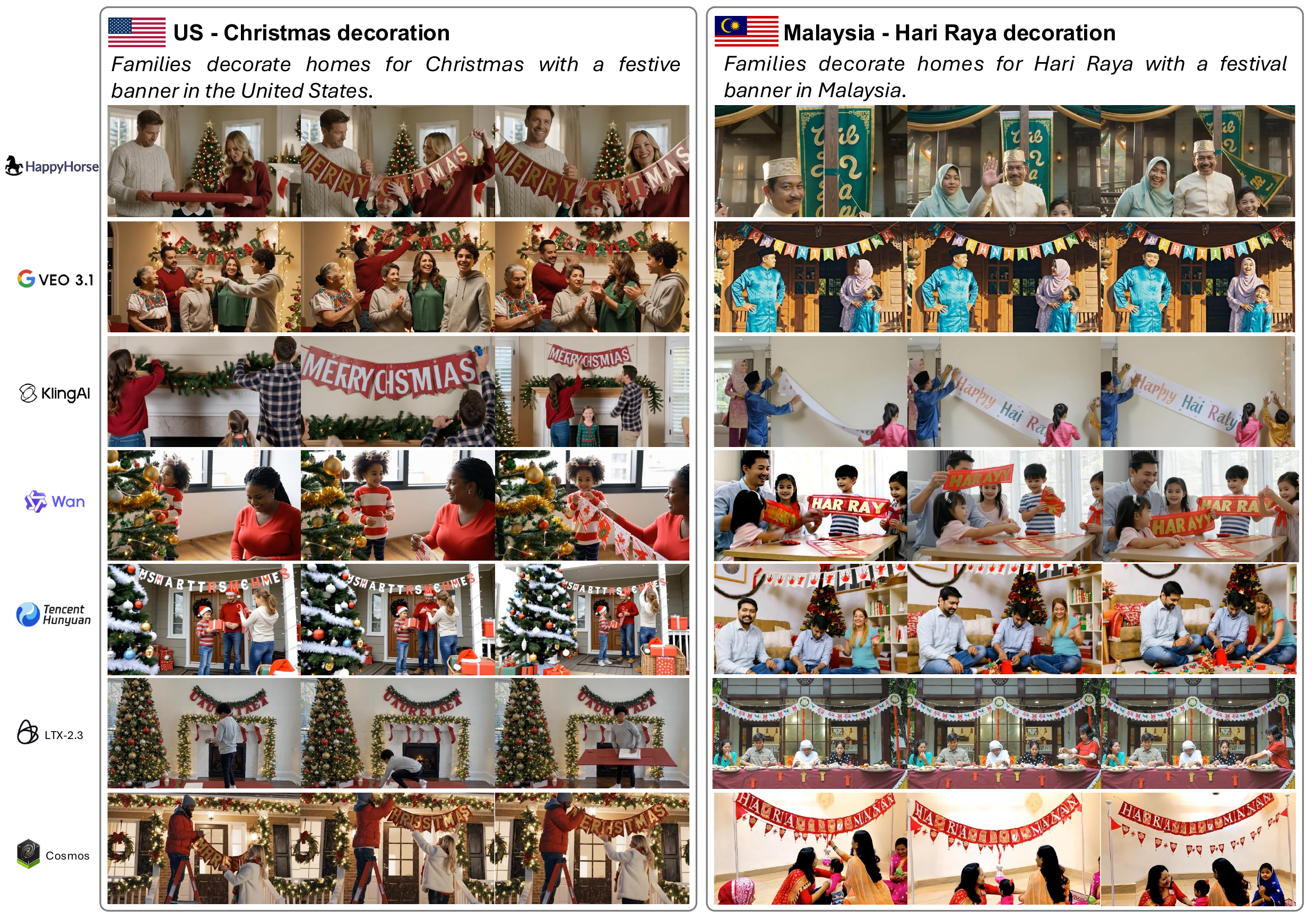}
	\caption{Sampled videos of different models with prompts from material culture category.}
	\label{Fig: case1}
\end{figure*}

\begin{figure*}[h]
	\centering
	\includegraphics[width=0.98\textwidth]{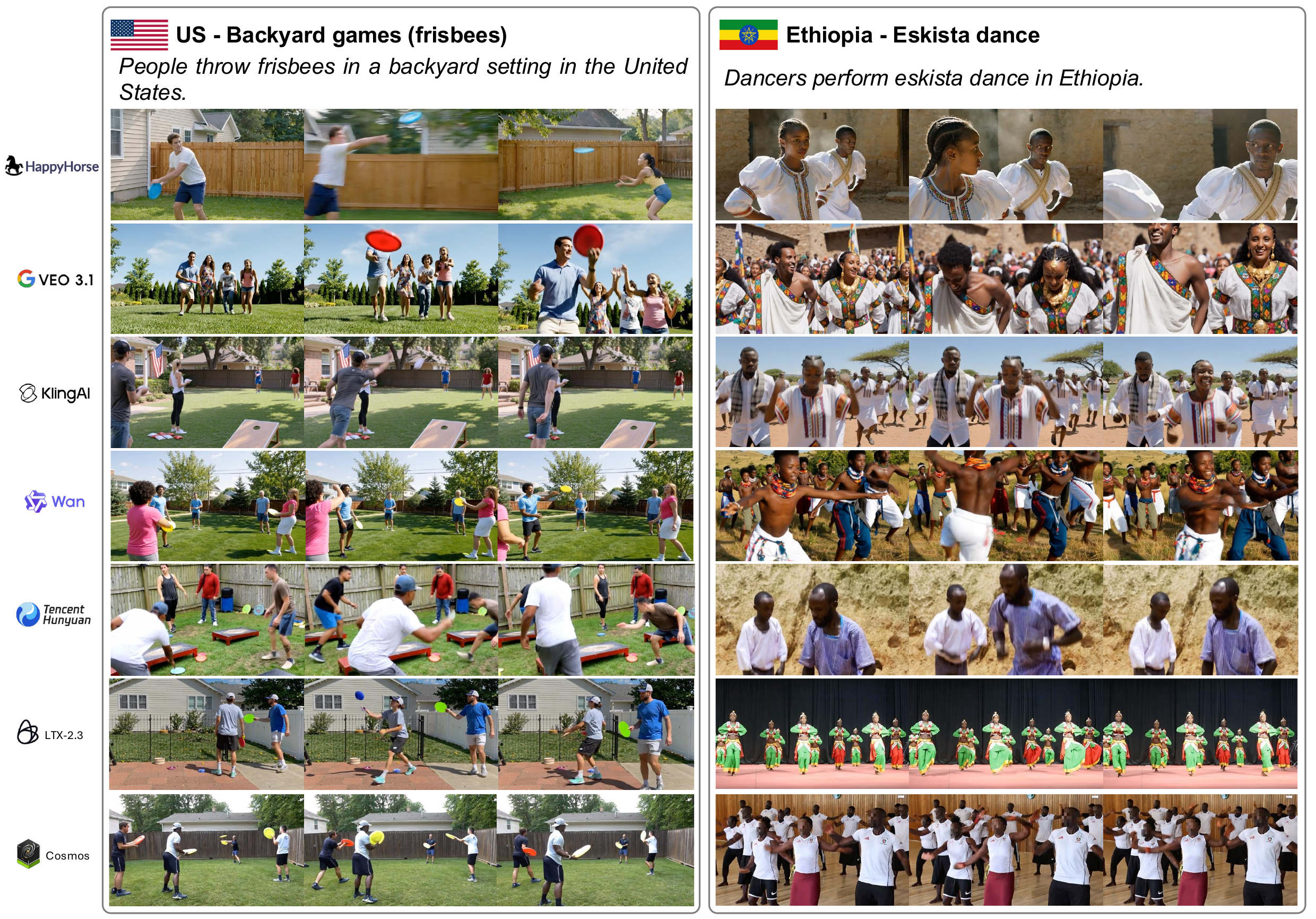}
	\caption{Sampled videos of different models with prompts from social practice \& performance category.}
	\label{Fig: case2}
\end{figure*}

\begin{figure*}[h]
	\centering
	\includegraphics[width=0.98\textwidth]{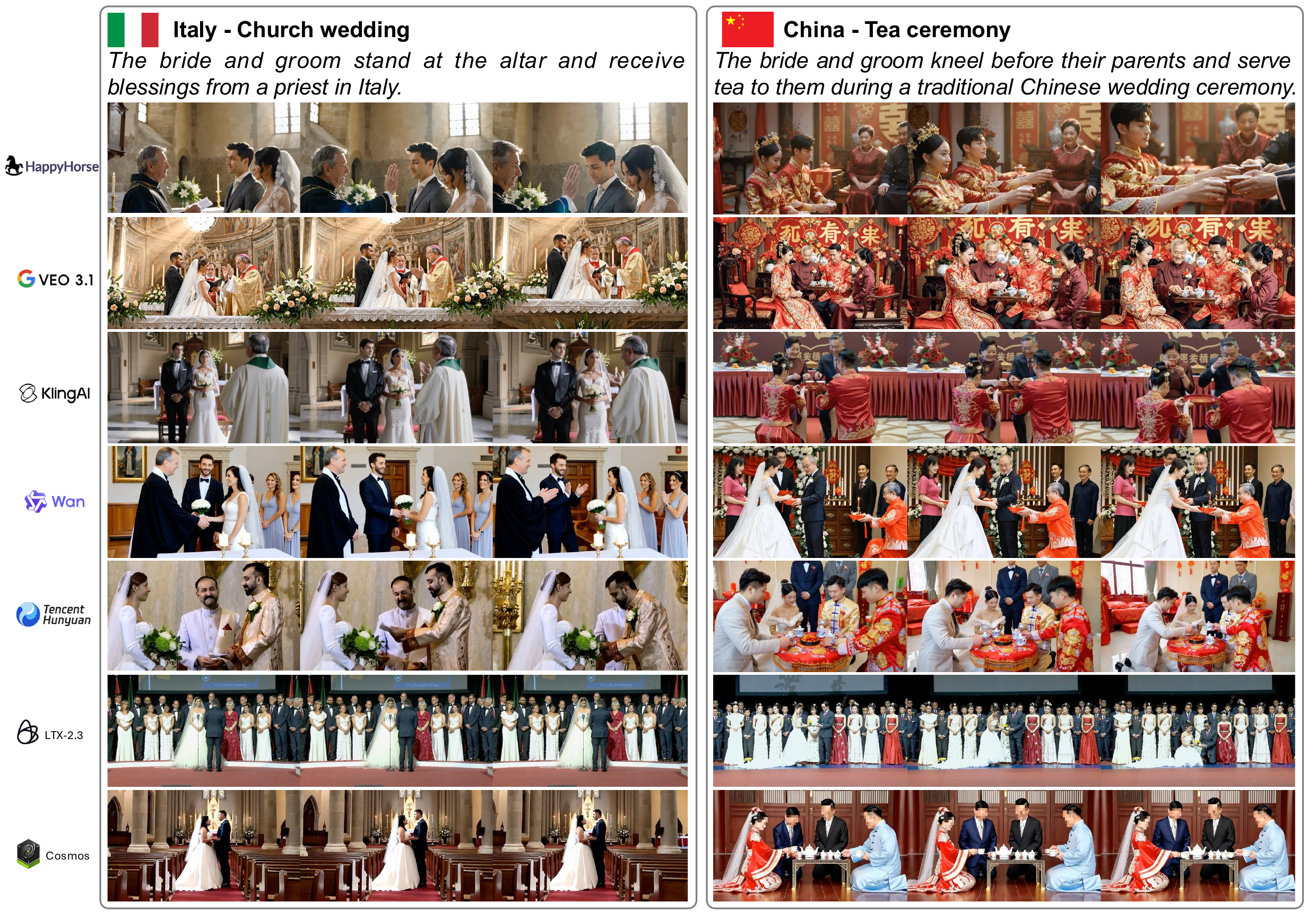}
	\caption{Sampled videos of different models with prompts from ritual \& ceremony category.}
	\label{Fig: case3}
\end{figure*}

\section{Qualitative Examples}
To provide a more intuitive analysis, we present qualitative examples across the three cultural categories in CultureVidBench: material culture, social practice \& performance, and ritual \& ceremony. For each category, we select representative prompts from different countries and provide video frames generated by seven T2V models.

\textbf{Material Culture.} For this category, we select two decoration-related cases: US-Christmas decoration and Malaysia-Hari Raya decoration, representing a high-resource cultural context and a relatively underrepresented cultural context, respectively. As shown in Figure~\ref{Fig: case1}, most models can capture the general visual appearance of Christmas decorations, such as Christmas trees, lights, and festive banners. However, despite good semantic adherence and visual quality, the generated banners often contain misspelled or unreadable text, showing that visible text rendering remains a major limitation for current T2V models.

The Hari Raya case is more challenging. Hari Raya is one of the most important and widely celebrated festivals in Malaysia, yet only a few models roughly capture its decorative style, such as HappyHorse-1.0. In contrast, several models generate scenes resembling Christmas decorations, such as HunyuanVideo-1.5. This indicates a bias toward high-resource cultural patterns and weaker understanding of underrepresented cultural contexts. Text rendering also remains unsuccessful in all the videos in this case, further highlighting the limitation of current models in generating culturally appropriate visible text.

\textbf{Social Practice \& Performance.} For this category, we select two cases: US-Backyard games with frisbees and Ethiopia-Eskista dance. The frisbee case involves relatively fast and large body motions, making it challenging for video generation. As shown in Figure~\ref{Fig: case2}, most T2V models can capture the general concept of a casual backyard scene with frisbees. However, videos generated by HunyuanVideo-1.5, LTX-2.3-22B, and Cosmos-Predict-2.5-14B contain noticeable visual artifacts or less realistic motion. These results reflect the performance gap between proprietary and open-source models in generating dynamic social activities.

In contrast, the Ethiopia-Eskista dance case receives consistently low human scores across models. Most models generate generic Ethiopian dance scenes but fail to capture the distinctive Eskista movement, especially the rapid shoulder shaking. This indicates that current T2V models may recognize the broad cultural context, such as Ethiopian performers and dance settings, but still fail to faithfully represent the specific cultural practice. The low human scores further confirm that this type of social performance remains challenging, as its cultural meaning depends on fine-grained body motion and performance style.

\textbf{Ritual \& Ceremony.}
For this category, we select two wedding-related cases: Italy-Church wedding and China-Tea ceremony. The Italy-Church wedding case requires the correct depiction of key ritual participants and interactions, including the bride and groom standing at the church altar and receiving blessings from a priest. As shown in Figure~\ref{Fig: case3}, in the Italy-Church wedding case, several models, including HappyHorse-1.0, Veo-3.1-Fast, Kling-3.0, HunyuanVideo-1.5, and Cosmos-Predict-2.5-14B, reasonably capture the cultural scene, such as the bride and groom standing in a church and receiving blessings from a priest. However, Wan-2.2-14B generates a more generic indoor wedding ceremony with a weaker sense of the church altar, while LTX-2.3-22B depicts a stage-like scene with rows of people, where the altar is not clearly presented. These results suggest that most models can recognize the overall ritual context, but some still lack fine-grained cultural details in representing religious wedding settings.

The China-Tea ceremony case requires the correct depiction of ritual interactions, such as kneeling before parents and serving tea to them, which involves understanding culturally specific procedures and participant relationships. HappyHorse-1.0 generates a culturally and semantically aligned video with a clear tea-serving action, an appropriate wedding setting, and correct visible text. Veo-3.1-Fast and Kling-3.0 also roughly capture the tea-serving procedure, but their visible text rendering remains inaccurate. However, Wan-2.2-14B, HunyuanVideo-1.5, LTX-2.3-22B, and Cosmos-Predict-2.5-14B show weaker performance, with incorrect procedural steps and culturally inconsistent details, such as Western-style wedding attire. This shows that ritual generation requires not only visual realism, but also an understanding of culturally specific actions, social roles, and procedural order. Such cultural expressions are also susceptible to interference from other cultural patterns, making it difficult for models to maintain cultural consistency throughout the generated video.

%Overall, these examples show that current T2V models remain limited in culturally grounded generation. Their limitations are particularly evident in underrepresented cultural contexts, where models often fail to capture fine-grained cultural details and are easily influenced by more dominant cultural patterns. This cross-cultural interference makes it difficult for models to maintain cultural consistency throughout the generated videos.

\section{Failure Analysis}
Overall, current T2V models remain limited in culturally grounded generation, particularly for underrepresented cultural contexts, rituals, and multimodal cultural rendering.

\textbf{Underrepresented cultural contexts.} 
Models often omit fine-grained local details and default to more dominant cultural patterns, potentially reflecting imbalanced cultural coverage in their training data. More geographically and linguistically balanced data, together with culture-aware data sampling, may improve representation of underrepresented cultures.

\textbf{Rituals.} Material culture is often associated with distinctive visual patterns that can be learned from repeated examples. In contrast, rituals require models to coordinate culturally specific actions, temporal order, and relationships among participants. These failures may therefore arise from both insufficient cultural knowledge and limited action planning. Procedural annotations, temporally ordered supervision, and explicit action planning may help address these limitations.

\textbf{Multimodal cultural rendering.} Text rendering requires exact symbolic structures rather than approximate visual resemblance, while existing models receive limited glyph-level supervision. Videos must additionally preserve character identity and location across frames, often resulting in distorted or flickering text. Script-aware rendering, OCR-based supervision, and temporal consistency constraints may help.

\end{document}